\documentclass[letterpaper]{article} 
\usepackage{aaai2026}  
\usepackage{times}  
\usepackage{helvet}  
\usepackage{courier}  
\usepackage[hyphens]{url}  
\usepackage{graphicx} 
\usepackage{amssymb}
\usepackage[numbers,square,comma,sort&compress]{natbib} 
\usepackage{caption} 
\usepackage{booktabs}
\usepackage{multirow}
\usepackage{amsmath}
\usepackage{amssymb}
\usepackage{algorithm}
\usepackage{algorithmic}
\usepackage{makecell}
\usepackage{xcolor}
\usepackage{newfloat}
\usepackage{listings}
\DeclareCaptionStyle{ruled}{labelfont=normalfont,labelsep=colon,strut=off} 
\floatstyle{ruled}
\newfloat{listing}{tb}{lst}{}
\floatname{listing}{Listing}
\title{INTENT: Invariance and Discrimination-aware Noise Mitigation for Robust Composed Image Retrieval
}
\author{
    Zhiwei Chen\textsuperscript{\rm 1},
    Yupeng Hu\textsuperscript{\rm 1}\thanks{Corresponding author.},
    Zhiheng Fu\textsuperscript{\rm 1},
    Zixu Li\textsuperscript{\rm 1},
    Jiale Huang\textsuperscript{\rm 1},
    Qinlei Huang\textsuperscript{\rm 1},
    Yinwei Wei\textsuperscript{\rm 1},
}
\affiliations{
    \textsuperscript{\rm 1}School of Software, Shandong University\\
    \{zivczw, fuzhiheng8, lizixu.cs\}@gmail.com, 
    \{huangjiale359, hql\}@mail.sdu.edu.cn, \\
    huyupeng@sdu.edu.cn,\
    weiyinwei@hotmail.com
}

\usepackage{bibentry}

\begin{document}
\maketitle

\begin{abstract}
Composed Image Retrieval (CIR) is a challenging image retrieval paradigm that enables to retrieve target images based on multimodal queries consisting of reference images and modification texts. Although substantial progress has been made in recent years, existing methods assume that all samples are correctly matched. However, in real-world scenarios, due to high triplet annotation costs, CIR datasets inevitably contain annotation errors, resulting in incorrectly matched triplets.
To address this issue, the problem of Noisy Triplet Correspondence (NTC) has attracted growing attention. We argue that noise in CIR can be categorized into two types: \textbf{cross-modal correspondence noise} and \textbf{modality-inherent noise}. The former arises from mismatches across modalities, whereas the latter originates from intra-modal background interference or visual factors irrelevant to the coarse-grained modification annotations. However, modality-inherent noise is often overlooked, and research on cross-modal correspondence noise remains nascent.
To tackle above issues, we propose the \textbf{I}nvariance and discrimi\textbf{N}a\textbf{T}ion-awar\textbf{E} \textbf{N}oise ne\textbf{T}work (\textbf{INTENT}), comprising two components: \textit{Visual Invariant Composition} and \textit{Bi-Objective Discriminative Learning}, specifically designed to handle the two-aspect noise. The former applies causal intervention on the visual side via Fast Fourier Transform (FFT) to generate intervened composed features, enforcing visual invariance and enabling the model to ignore modality-inherent noise during composition.
The latter adopts collaborative optimization with both positive and negative samples, and constructs a scalable decision boundary that dynamically adjusts decisions based on the loyalty degree, enabling robust correspondence discrimination.
Extensive experiments on two widely used benchmark datasets demonstrate the superiority and robustness of INTENT. Codes are available at \url{https://github.com/zivchen-ty/INTENT/}

\end{abstract}

\begin{figure}[t]
\begin{center}
\includegraphics[width=\linewidth]{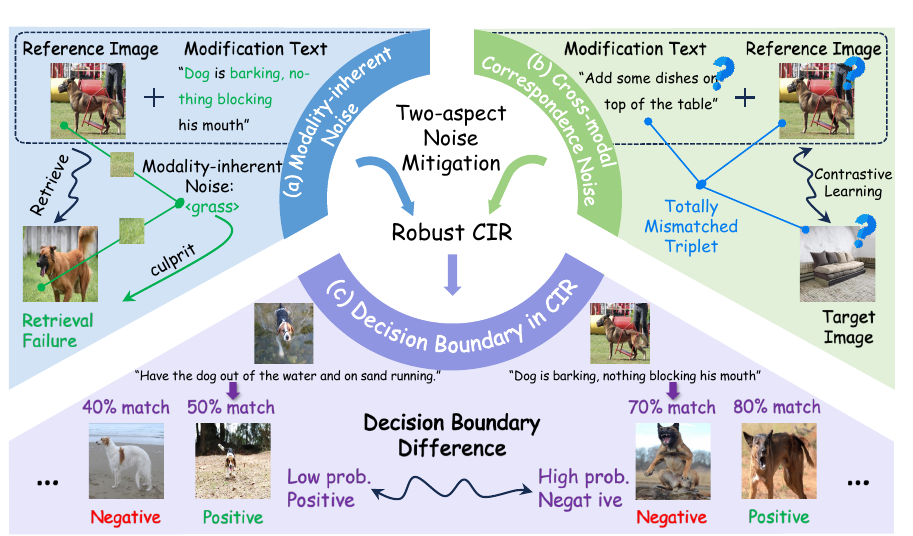}
\end{center}
   \caption{(a) shows typical Modality-inherent Noise in CIR. (b) reveals Cross-modal Correspondence Noise. (c) presents two cases: left successfully retrieves target with low confidence (50\%), while right fails with high confidence (70\%), illustrating varying decision boundaries in retrievals.}
\label{fig:intro}
\end{figure}
\section{Introduction}\label{sec:intro}
In recent years, the rapid growth of multimedia data~\cite{wang2024twin,ge2025debate,wang2025r1trackdirectapplicationmllms,ni2025wonderturbo,ni2025recondreamerRL,ni2025recondreamer} has brought increasing attention to Composed Image Retrieval (CIR)~\cite{sprc, limn, ssn, HABIT, HINT, MELT}. Unlike traditional single-modal or cross-modal retrieval~\cite{targetrecognition,MAIL}, CIR enables more flexible retrieval through multimodal queries, specifically by combining a reference image and a modification text, to accurately retrieve a target image that satisfies specific requirements~\cite{ReTrack,REFINE,STABLE}. This paradigm has shown significant value in various domains, such as information processing~\cite{lan2025multi,liu2024dvlo,jiang2025transforming,bi2025cot,zhou2024information,qiu2024tfb,Feng2023MaskCon,anonymous2025comptrack,yuan2025deep,qiu2025duet,BML,zhang2026decoding,wang2026eeo,zhou2024lidarptq}, intelligent systems~\cite{wen2023syreanet,lu2025does,zhou2025dragflow,wang2024computing,yu2025visualizing,bi2025llava,li2023hong,duan2025copinn,yu2025iidm}, and multimodal learning~\cite{cheng2026enhancing,chen2025autoneural,jia2026ramrecover3dhuman,li2025human,li2026multiple,ye2026gigaworld,ERASE}, effectively addressing users' diverse retrieval needs.

Although existing CIR methods have made significant progress, most of them tend to overlook the issue of Noisy Triplet Correspondence (NTC) in query data, i.e., possible incorrectly
matched triplets. Specifically, as illustrated in Figure~\ref{fig:intro}(a) and (b), NTC in CIR is primarily caused by the following two factors: 
\textbf{modality-inherent noise} and \textbf{cross-modal correspondence noise}. The former is introduced by intra-modal interference factors, such as complex backgrounds in the image or visual elements irrelevant to the modification requirements, while the latter arises from incorrect semantic alignment across modalities. As research on NTC in CIR remains at an early stage, achieving robust CIR learning presents two major challenges.

\textbf{C1: Neglected inherent noise.} Existing studies on noise in CIR primarily focus on addressing cross-modal correspondence noise~\cite{TME}, with limited analysis or solutions for modality-inherent noise. Such methods typically perform direct fusion of multimodal query features and determine noisy correspondences based on the similarity between composed and the target features. However, they often overlook the fact that modality-inherent noise, such as interference from irrelevant content in the reference image, may distort the composition process and lead to inaccurate feature fusion, thereby introducing uncertainty in noisy correspondence identification.
While several recent CIR models attempt to handle modality-inherent noise~\cite{dwc,encoder}, they generally assume that cross-modal correspondence is correct and rely on cross-modal interaction to suppress the intra-modal noisy signals. 
In the context of NTC, where incorrect cross-modal matching may occur, these methods are difficult to apply directly. Therefore, achieving effective noise mitigation under uncertain matching conditions poses the first major challenge in robust CIR learning.

\textbf{C2: Hard decision boundary.} Recent work have focused on noisy dual correspondence (NDC) problem and proposed various strategies such as adversarial training~\cite{ndc-1, ndc-2}, sample reweighting~\cite{ndc-3}, and self-supervised learning~\cite{ndc-5, ndc-6}. However, these methods are mainly designed for noise discrimination in paired data and are not readily applicable to the Noisy Triplet Correspondence (NTC) problem. Moreover, current NTC approaches~\cite{TME} typically employ a hard decision boundary.
Due to the brevity of modification text in CIR, capturing the semantic gap between reference and target images is challenging.
Consequently, similarity scores don't absolutely correspond to noise presence, especially with varying visual complexity across samples.
As shown in Figure~\ref{fig:intro}(c), a 50\% similarity may suffice for positive labels in some cases (i.e., the left one), while even a 70\% score may be negative if the content is simple or there are better-matching candidates in the dataset (i.e., the right one).
Thus, accurate noise discrimination in CIR requires contrastive comparison between positive and negative samples, and the design of a scalable decision boundary for robust composed feature alignment, constitutes the second major challenge.

To address these challenges, we propose the Invariance and discrimiNaTion-awarE Noise mitigation neTwork (INTENT), for handling both modality-inherent noise and ambiguous decision boundaries in Composed Image Retrieval (CIR). INTENT comprises two main modules:
\textit{(1) The Visual Invariant Composition (VIC) module} applies causal intervention on reference image via Fast Fourier Transform (FFT) to generate counterfactual sample, utilizing it to further enforce visual invariance and enabling the model to ignore modality-inherent noise during composition.
\textit{(2) The Bi-Objective Discriminative Learning (BiODL) module} performs collaborative optimization using positive and negative samples, constructing a scalable decision boundary that dynamically adjusts decisions based on the loyalty degree of sample matching, enabling more robust composed feature alignment and improving model's discrimination.

The main contributions of this paper are as follows:
\begin{itemize}
    \item We are the first to clearly distinguish and investigate two-aspect noise which may lead to Noisy Triplet Correspondence (NTC), focusing on the challenges of neglected inherent noise and hard decision boundaries, and revealing their critical impact on model accuracy and robustness.
    \item We propose INTENT, a novel robust CIR framework. The VIC module mitigates modality-inherent noise via causal intervention, while the BiODL module performs collaborative optimization over positive and negative samples and constructs a scalable decision boundary for more robust composed feature alignment.
    \item Extensive experiments on multiple benchmarks show that INTENT significantly outperforms most methods in both accuracy and robustness, confirming the effectiveness of our approach.
\end{itemize}

\section{Related Work}
Our work is closely related to Composed Image Retrieval (CIR) with Noisy Correspondence and Causal Intervention.

\noindent\textbf{Composed Image Retrieval with Noisy Correspondence}.
Composed Image Retrieval (CIR) retrieve target images using reference images combined with modification texts. Current approaches follow two paradigms. Early works~\cite{tirg, clvcnet} used conventional architectures like ResNet and LSTM for separate feature extraction before fusion. Recent advances leverage pre-trained vision-language models such as CLIP~\cite{clip} for joint learning, achieving superior results through streamlined alignment and composition~\cite{xie2026conquer,xie2026hvd,gu2025mocount,li2025chatmotion,jia2024adaptive,liu2024graph}. 
While most CIR research assumes well-aligned triplet annotations, large-scale datasets often contain noisy triplet correspondence (NTC) due to annotation errors or semantic ambiguity~\cite{xie2026delving,xie2025chat,song5,li2024synergized,zeng2025bridging,cao2026task}. Addressing such robust learning~\cite{sun2025roll,Feng2022SSR,he2024robust,Feng2024NoiseBox,sun2024robust,li2025set,xu2025noisy,Feng2024CLIPCleaner} of noisy correspondences is challenging, as it requires distinguishing reliable supervision from unreliable ones during training. Recent work has begun to tackle this problem through sample selection and realignment strategies, making CIR models more robust to annotation noise~\cite{TME}. 
While some methods~\cite{mgur} address false positives, they mainly aim to improve CIR performance rather than addressing robustness under NTC, thus not suited for tackling challenges posed by noisy triplet correspondence.

\noindent\textbf{Causal Intervention in Multimodal Learning.}
Causal intervention techniques have attracted increasing attention in multimodal learning~\cite{song3,meng2026tri,wang2025ascd,gao2024eraseanything,yu2025visual,song4,bi2025prismselfpruningintrinsicselection,ge2025beyond,song13,zhang2026towards}, especially for computer vision~\cite{song1,liao2024globalpointer,zhang2023multi,song2,liu2023regformer,yu2025yielding,zhou2023fastpillars,liao2025convex,pillarhist,focustrack,zhao2024balf,liu2024difflow3d,song6,liu2025difflow3d,zhang2024cf,jiang2025stg,yu2025cotextor}, visual question answering~\cite{counterfactual-vqa,wang2025see,ge2024consistencies,ge2025gen4track}, and so on. 
Typical approaches in this domain, such as backdoor adjustment and counterfactual data augmentation, are primarily designed to mitigate the adverse effects of spurious correlations. They achieve this by simulating structural interventions on specific input variables to sever non-causal links. By doing so, these methods compel models to prioritize genuine causal structures, thereby significantly enhancing their robustness and generalization capabilities against unseen confounders or irrelevant environmental variations. Recently, intervention-based methodologies have been successfully adapted for cross-modal tasks, particularly in image-text scenarios~\cite{dcin}. By performing interventions at the feature level, these models can effectively decouple entangled attributes, leading to representations that are more strictly aligned with true semantic relationships rather than dataset-specific biases. Despite these promising advances, the application of causal intervention in more intricate multimodal scenarios—most notably, composed image retrieval (CIR)—remains largely underexplored. CIR inherently requires the delicate synthesis of visual and textual cues, making it particularly susceptible to complex compounding confounders. To bridge this gap, we introduce novel, intervention-inspired strategies explicitly tailored for CIR, effectively encouraging a more robust, generalizable, and causally meaningful composition of image and text features.

\begin{figure}[t]
\begin{center}
\includegraphics[width=\linewidth]{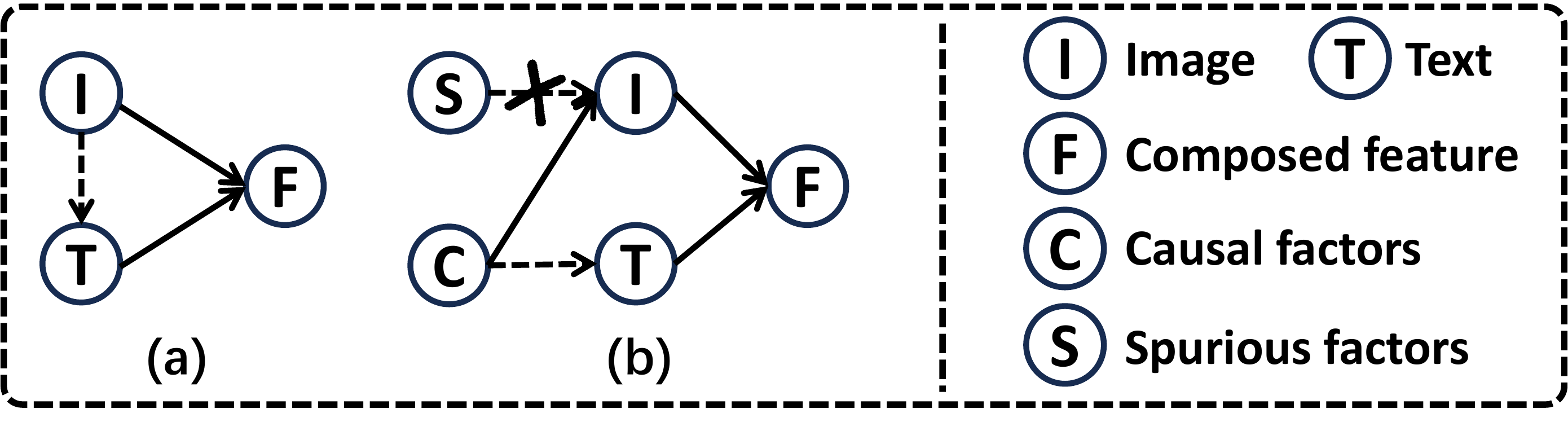}
\end{center}
   \caption{The causal graph of CIR. Solid arrows present the \textbf{cause effect}. Dash arrows mean there exist \textbf{correlations}.}
\label{fig:causal}
\end{figure}

\begin{figure*}
\begin{center}
\includegraphics[width=1.0\linewidth]{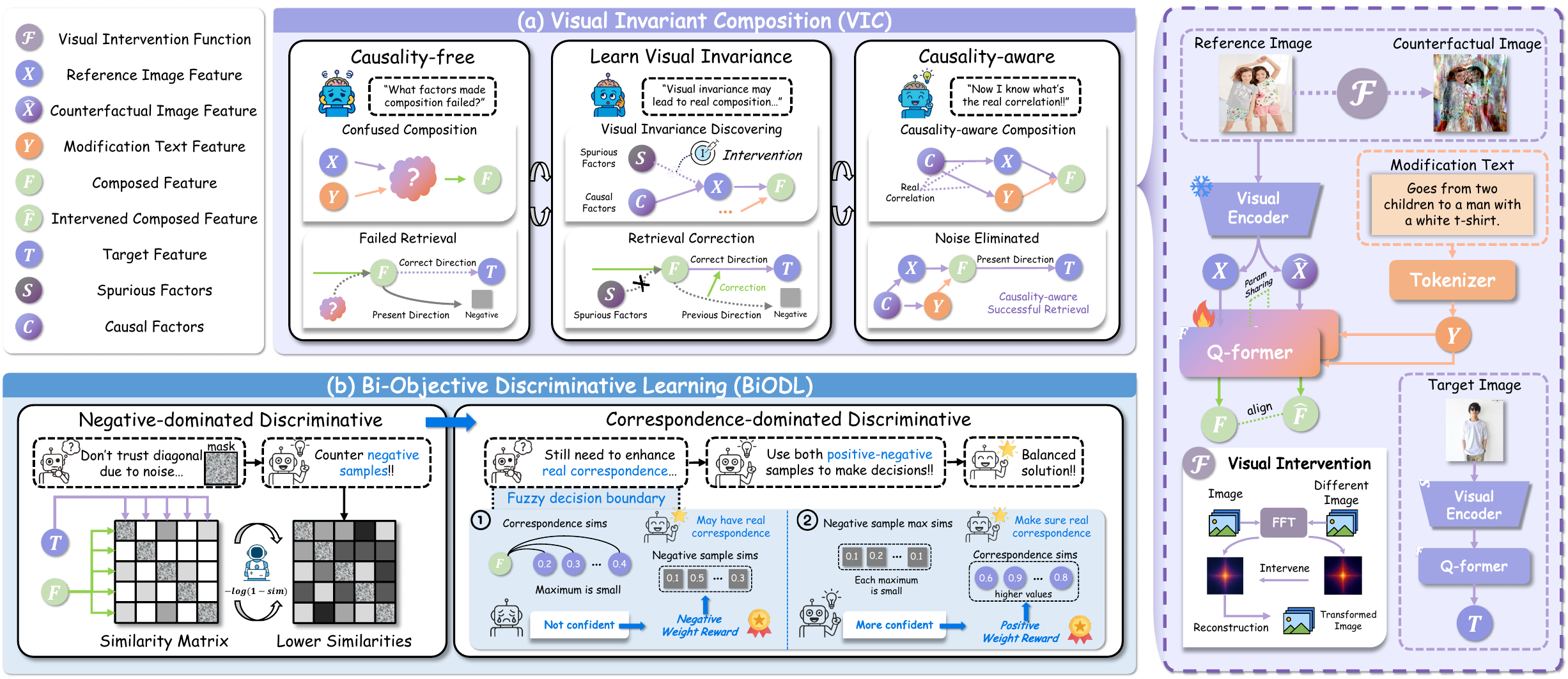}
\end{center}
   \caption{The framework of our proposed INTENT. We designed (a) Visual Invariant Composition and (b) Bi-Objective Discriminative Learning, to mitigate \textbf{modality-inherent noise} and \textbf{cross-modal correspondence noise}, respectively.}
\label{fig:overview}
\end{figure*}

\section{The Proposed INTENT}
In this section, we introduce INTENT, as Figure~\ref{fig:overview} shows.

\subsection{CIR from a Causal Perspective}
\label{sec:3.1}
\noindent\textbf{Task definition.} Given a set of $N$ multimodal queries $\mathcal{Q}=\left\{\left(x_{r},y_{m}\right)_{n}\right\}_{n=1}^{N}$, where $x_{r}, y_{m}$ refer to the reference image and modification text, respectively. CIR aims to retrieve the most relevant target image $t$ for each query $\mathcal{Q}$. 

\noindent\textbf{Causal analysis.} 
In CIR tasks, since modification texts typically specify changes to partial content in reference images, they naturally share semantic correlations. Therefore, to eliminate modality-inherent noise, we consider leveraging these semantic correlations during multimodal composition to identify noise content.
As shown in Figure~\ref{fig:causal}(a), the CIR composition process can be modeled as $I\longrightarrow F\longleftarrow T$. Additionally, the specific content of reference image $I$ influences the interpretation of modification instructions in text $T$, creating $I\dashrightarrow T$.
So if we could model the correlation between $I$ and $T$, we could explicitly guide the composition. However, this correlation is unobservable due to lack of supervision signals. Thus we explore causal relations during composition to achieve indirect understanding of these correlations.
As Figure~\ref{fig:causal}(b) shows, reference image $I$ comprises causal factors $C$ and spurious factors $S$. Causal factors $C$ represent true semantic attributes the text intends to modify, reflecting genuine causal relations. Spurious factors $S$ represent irrelevant, non-causal visual information (e.g., unmodified objects) that corrupts the composition process, i.e., modality-inherent noise.
To eliminate modality-inherent noise, we suppress spurious factors. Since modification texts are user-written and concise, we consider modality-inherent noise primarily exists in reference images.

Two potential paths exist for suppressing spurious factors: 1) precisely decouple image features and remove spurious factors, 2) ignore possible spurious factors during composition. 
While the first approach is more direct, it's difficult to implement. Existing works in this direction focus on implicit operations~\cite{pair}, but uncontrollable implicit processes may cause inaccurate decoupling, harming training. Therefore, we choose the second path, ensuring the final composed features are less affected by modality-inherent noise, implemented by Visual Invariant Composition (VIC).

\subsection{Visual Invariant Composition}
\label{sec:3.2}
As shown in Figure~\ref{fig:overview}(a), the Visual Invariant Composition (VIC) performs an intervention described in Section~\ref{sec:3.1}, forcing the model to focus only on semantics invariant to spurious factors, thus ignoring modality-inherent noise in reference images.
This is achieved by enforcing consistency between two composed features derived from the reference image and its counterfactual, intervened version.
 
\noindent\textbf{Counterfactual Image Generation.}
To achieve this, we first perform a frequency-domain intervention on the reference image $x_{r}$, perturbing its original frequency components while preserving semantic structures. This process yields a counterfactual image $\hat{x}_r$ in which modality-inherent noise is altered, but the core semantic content remains unchanged.

Specifically, as shown in the right part of Figure~\ref{fig:overview}, for each reference image, we randomly sample an irrelevant image $x_{d}$, and apply Fast Fourier Transform (FFT), which which has been widely used in the vision domain for manipulating image signals, to both $x_r$ and $x_d$, obtaining their spectra $\textbf{F}_r=\mathcal{F}(x_r), \textbf{F}_d=\mathcal{F}(x_d)$, where $\mathcal{F}(\cdot)$ denotes the FFT function.
The spectrum combines an amplitude and a phase spectrum, where amplitude captures low-level statistics (style, texture) while phase contains high-level semantic structures. 
We intervene solely on amplitude spectra by randomly mixing the cropped central regions with random ratio $\lambda$, represented as follows, 
\begin{equation}
\hat{\textbf{A}}_r = \lambda \textbf{A}_d^{\text{crop}} + (1 - \lambda) \textbf{A}_r^{\text{crop}},
\end{equation}
where $\textbf{A}_r, \textbf{A}_d$ are amplitude spectra of $\textbf{F}_r, \textbf{F}_d$, $\textbf{A}^{\text{crop}}$ denotes the amplitude spectra of cropped central region.
Finally, we reconstruct the counterfactual image via inverse FFT: $\hat{x}_r = \mathcal{F}^{-1}(\hat{\textbf{A}}_r, \theta_r)$, where $\theta_r$ is the phase spectrum of $\textbf{F}_r$.

In this way, $\hat{x}_r$ preserves key semantics of the original $x_r$ while exhibiting altered noise patterns, providing ideal data pairs for subsequent consistency learning.

\noindent\textbf{Causality-free Composition.}
After obtaining the factual-counterfactual image pairs, we leverage BLIP-2's Q-Former architecture for multimodal feature composition, combining the reference image ${x}_r$ and counterfactual image $\hat{x}_r$, with the modification text respectively, formulated as,
\begin{equation}
\left\{
\begin{aligned}
    &\textbf{F}_c=\operatorname{Q-Former}(\varPhi_\mathbb{X}(x_r),\varPhi_\mathbb{Y}(y_m)),\\
    &\hat{\textbf{F}}_c=\operatorname{Q-Former}(\varPhi_\mathbb{X}(\hat{x}_r),\varPhi_\mathbb{Y}(y_m)),
\end{aligned}
\right.
\label{Q-Former}
\end{equation}
where $\textbf{F}_c, \hat{\textbf{F}}_c\in\mathbb{R}^{Q\times D}$ represent the composed feature and intervened composed feature. $Q$ is number of Q-former's learnable queries, and $D$ is the feature dimension. $\varPhi_\mathbb{X}$ and $\varPhi_\mathbb{Y}$ denote BLIP-2's frozen visual encoder and tokenizer respectively. Similarly, for target image $t$, we obtain its feature $\textbf{F}_t\!=\!\operatorname{Q-Former}(\varPhi_\mathbb{X}(t))\!\in\!\mathbb{R}^{Q\times D}$.
Notably, since the intervened composed feature have not been used for training, the model remains causality-free, unable to identify correct causal relations and still affected by spurious factors.

\begin{table*}[t]
\centering
\setlength{\tabcolsep}{4pt}
\begin{tabular}{c|l|cccc|ccc|c}
\hline

\multirow{2}{*}{\textbf{Noise}} & \multirow{2}{*}{\textbf{Methods}} 
& \multicolumn{4}{c|}{\textbf{R@K}} 
& \multicolumn{3}{c|}{\textbf{R$_{sub}$@K}} 
& \multirow{2}{*}{\textbf{Avg(R@5, R$_{sub}$@1)}} \\
\cline{3-9}
& & K=1 & K=5 & K=10 & K=50 & K=1 & K=2 & K=3 & \\
\hline
\hline
\multirow{7}{*}{0\%}
& SSN~\cite{ssn}~(AAAI'24) & 43.91 & 77.25 & 86.48 & 97.45 & 71.76 & 88.63 & 95.54 & 74.51 \\
          & CALA~\cite{cala}~(SIGIR'24) & 49.11 & 81.21 & 89.59 & 98.00 & 76.27 & 91.04 & 96.46 & 78.74 \\
          & SPRC~\cite{sprc}~(ICLR'24) & 51.96 & 82.12 & 89.74 & 97.69 & \underline{80.65} & 92.31 & 96.60 & 81.39 \\
         & RCL~\cite{RCL}~(TPAMI'23) & 53.16 & 82.41 & 90.12 & \textbf{98.34} & 79.57 & 92.02 & 96.87 & 80.99 \\
          & RDE~\cite{rde}~(CVPR'24) & 51.81 & 82.02 & \underline{90.60} & 97.93 & 78.17 & 91.90 & 96.70 & 80.10 \\
          & TME~\cite{TME}~(CVPR'25) & \textbf{53.42} & \underline{82.99} & 90.24 & 98.15 & \textbf{81.04} & \textbf{92.58} & \textbf{96.94} & \textbf{82.01} \\
          & \textbf{INTENT (Ours)} & \underline{53.37} & \textbf{83.16} & \textbf{90.73} & \underline{98.22} & {80.24} & \underline{92.37} & \underline{96.89} & \underline{81.70} \\
     \hline
\multirow{7}{*}{20\%}
& SSN~\cite{ssn}~(AAAI'24) & 34.02 & 65.90 & 75.78 & 91.33 & 66.92 & 85.90 & 93.45 & 66.41 \\
& CALA~\cite{cala}~(SIGIR'24) & 41.33 & 72.70 & 82.84 & 94.34 & 71.66 & 88.15 & 94.94 & 72.18 \\
& SPRC~\cite{sprc}~(ICLR'24) & 45.90 & 75.86 & 83.52 & 93.37 & \underline{78.10} & \underline{91.40} & 96.05 & 76.98 \\
& RCL~\cite{RCL}~(TPAMI'23) & 50.43 & \underline{81.11} & \underline{88.82} & 96.68 & 77.52 & 90.80 & 95.71 & 79.31 \\
& RDE~\cite{rde}~(CVPR'24) & 49.23 & 78.63 & 86.80 & 95.78 & 76.58 & 90.31 & 96.07 & 77.60 \\
& TME~\cite{TME}~(CVPR'25) & \textbf{51.35} & 81.01 & 88.53 & \underline{97.81} & \textbf{78.46} & 91.25 & \underline{96.39} & \textbf{79.74} \\
& \textbf{INTENT (Ours)} & \underline{51.25} & \textbf{81.36} & \textbf{90.02} & \textbf{98.05} & {77.95} & \textbf{91.40} & \textbf{96.46} & \underline{79.66} \\
\hline
\multirow{7}{*}{50\%}
& SSN~\cite{ssn}~(AAAI'24) & 25.93 & 53.71 & 63.40 & 82.10 & 62.10 & 82.27 & 91.57 & 57.90 \\
& CALA~\cite{cala}~(SIGIR'24) & 36.10 & 66.12 & 77.76 & 92.10 & 68.12 & 85.66 & 93.59 & 67.12 \\
& SPRC~\cite{sprc}~(ICLR'24) & 39.93 & 66.00 & 73.59 & 86.48 & 75.81 & 89.21 & 95.37 & 70.90 \\
& RCL~\cite{RCL}~(TPAMI'23)& \underline{48.58} & 77.45 & 85.93 & 94.70 & 75.60 & 89.28 & 94.80 & 76.52 \\
& RDE~\cite{rde}~(CVPR'24) & 45.98 & 75.30 & 83.73 & 94.48 & 73.98 & 88.99 & 95.13 & 74.64 \\
& TME~\cite{TME}~(CVPR'25) & 48.48 & \underline{78.94} & \underline{87.28} & \underline{96.99} & \underline{76.48} & \underline{90.07} & \underline{95.83} & \underline{77.71} \\

& \textbf{INTENT (Ours)} & \textbf{49.78} & \textbf{79.64} & \textbf{88.99} & \textbf{97.37} & \textbf{77.18} & \textbf{90.41} & \textbf{96.00} & \textbf{78.41} \\
\hline
\multirow{7}{*}{80\%}
& SSN~\cite{ssn}~(AAAI'24) & 20.48 & 43.98 & 54.27 & 74.80 & 56.48 & 77.20 & 89.54 & 50.23 \\
& CALA~\cite{cala}~(SIGIR'24) & 31.52 & 61.49 & 72.60 & 89.86 & 64.34 & 83.52 & 92.60 & 62.92 \\
& SPRC~\cite{sprc}~(ICLR'24) & 29.95 & 51.25 & 58.51 & 73.86 & 70.22 & 86.05 & 93.21 & 60.74 \\
& RCL~\cite{RCL}~(TPAMI'23) & 44.94 & 74.43 & 82.99 & 92.31 & 71.93 & 86.84 & 92.96 & 73.18 \\
& RDE~\cite{rde}~(CVPR'24) & 42.92 & 71.30 & 80.51 & 92.96 & 69.64 & 85.86 & 93.54 & 70.47 \\
& TME~\cite{TME}~(CVPR'25) & \underline{46.31} & \underline{75.78} & \underline{84.89} & \underline{95.83} & \underline{73.37} & \underline{88.02} & \underline{94.89} & \underline{74.58} \\
& \textbf{INTENT (Ours)} & \textbf{47.90} & \textbf{78.13} & \textbf{87.04} & \textbf{96.47} & \textbf{73.81} & \textbf{89.18} & \textbf{95.54} & \textbf{75.97} \\
\hline
\end{tabular}
\caption{Performance comparison on the CIRR test set in terms of R@K(\%) and R$_{sub}$@K(\%). The best and second-best results are highlighted in \textbf{bold} and \underline{underline}, respectively.} 
\label{tab:cirr-noise}
\end{table*}

\noindent\textbf{Visual Invariance Learning.}
In the previous stage, we utilized reference and counterfactual images for composition respectively, obtaining composed features $\textbf{F}_c$ and intervened composed feature $\hat{\textbf{F}}_c$ that inherit different modality-inherent noise. 
Building upon this, we only need to train the model to identify visual invariants at the feature level, thereby homogenizing noise across different reference images, reducing the model's sensitivity to noise, and indirectly achieving modality-inherent noise mitigation.

To this end, we design visual invariance learning, which forces the model to maintain understanding of key visual information under different modality-inherent noise, obtaining unbiased composed features.
Specifically, we define a Causal Consistency Loss based on Centered Kernel Alignment (CKA)~\cite{cka_loss}, to measure the semantic consistency between $\mathbf{F}_c$ and $\hat{\mathbf{F}}_c$. We first compute the Gram matrices $\mathbf{K}_c\!\! = \!\!\mathbf{F}_c \mathbf{F}_c^{\top}\in\mathbb{R}^{Q\times Q}$ and $ \quad \!\!\!\!\!\!\mathbf{L}_c\!\! =\!\! \hat{\mathbf{F}}_c \hat{\mathbf{F}}_c^{\top}\in\mathbb{R}^{Q\times Q}$. Then we perform centering to eliminate mean shift:
\begin{equation}
\bar{\mathbf{K}}_c = \mathbf{H} \mathbf{K}_c \mathbf{H}, \quad \bar{\mathbf{L}}_c = \mathbf{H} \mathbf{L}_c \mathbf{H},
\end{equation}
where $\bar{\mathbf{K}}_c, \bar{\mathbf{L}}_c\in\mathbb{R}^{Q\times Q}$, $\mathbf{H} = \mathbf{I} - \frac{1}{Q} \mathbf{e}\mathbf{e}^{\top}$, $\mathbf{I}$ is the identity matrix, $\mathbf{e}$ is the unit vector, and $Q$ is the number of Q-former's learnable queries.
Finally, we compute the CKA similarity as a causal consistency constraint:
\begin{equation}
\mathcal{L}_{\text{caco}} = \frac{1}{B}\sum_{i=1}^{B}(1 - \frac{\langle \bar{\mathbf{K}}_{ci}, \bar{\mathbf{L}}_{ci} \rangle_F}{\|\bar{\mathbf{K}}_{ci}\|_F \, \|\bar{\mathbf{L}}_{ci}\|_F}),
\label{cka}
\end{equation}
where $B$ represents the batch size, $\langle \cdot, \cdot \rangle_F$ denotes Frobenius inner product,$\|\cdot\|_F$ is the Frobenius norm.
$\bar{\mathbf{K}}_{ci}$ and $\bar{\mathbf{L}}_{ci}$ represent the $i$-th Gram matrix of composed features in a batch. 
By this, the model becomes causality-aware, capable of mining visual invariants while ignoring modality-inherent noise, providing a causal correction for composed features.

\subsection{Bi-Objective Discriminative Learning}
\label{sec:3.3}
While mitigating modality-inherent noise, to enhance the model's generalization capability in NTC problem, we propose Bi-Objective Discriminative Learning (BiODL), which jointly optimizes decision boundaries from both negative-dominated and correspondence-dominated perspectives.

\noindent\textbf{Negative-dominated Discriminative Learning.}
Traditional CIR methods often use InfoNCE~\cite{tirg} to pull positive samples closer and push negative samples apart. However, in NTC scenarios, positive sample pairs may contain noisy matches, leading to unreliability. Therefore, inspired by RCL~\cite{RCL}, we design a Robust Contrastive Loss that focuses on reducing the negative impact of negative samples, formulated as:
\begin{equation}
\mathcal{L}_{\text{robust}} = -\frac{1}{B} \sum_{i,j\neq{i}}^{B} \log (1-\frac{\exp \left\{ \mathbf{F}_{ci}\mathbf{F}_{ti}^{\top}  / \tau\right\}}{ \sum_{j=1}^{B} \exp \left\{ \mathbf{F}_{ci}\mathbf{F}_{tj}^{\top} / \tau \right\}  } ),
\label{robust}
\end{equation}
where $B$ is batch size, $\tau$ is the temperature factor. $\mathbf{F}_{ci}$ and $\mathbf{F}_{tj}$ are $i$-th query feature and $j$-th target feature respectively.

The process actively pushes away negative samples, achieving spatial dispersion between positive and negative samples. Notably, although there may be correct correspondences among negative samples, the harm of treating them as negatives for learning is far less than treating noisy correspondences as positives, thus being more robust.

\noindent\textbf{Correspondence-dominated Discriminative Learning.}
While noisy correspondence interference can be reduced at negative-dominated level, the model requires active enhancement of discrimination among reliable correspondence.
Thus we propose a scalable decision boundary based on loyalty degree.
First, we define the similarity matrix between queries and targets $\mathbf{S} = 
\left\{
    \textbf{s}_{ij}| \textbf{s}_{ij}=\operatorname{softmax}(\operatorname{s} \left(\mathbf{F}_{ci}^{\top}\mathbf{F}_{tj}\right))
\right\}$,
where $\operatorname{s}(\cdot, \cdot)$ denotes similarity computation, $\mathbf{F}_{ci}$, $\mathbf{F}_{tj}$ $\in\mathbb{R}^{1\times D}$ represent the $i$-th composed feature and the $j$-th target feature in a mini-batch.

Based on the similarity matrix, we define the maximum positive matching likelihood $p^+$, and maximum negative matching likelihood $p^-$ for each query to subsequently estimate matching loyalty degree:
\begin{equation}
p_i^{+} = \max_{j}(\textbf{s}_{ij}\cdot y_{ij}), \quad p_i^{-} = \max_{j}(\textbf{s}_{ij}\cdot(1 - y_{ij})),
\end{equation}
where $y_{ij}$ denotes the pseudo-label matrix (diagonal elements are 1, others are 0), $\textbf{s}_{ij}$ denotes the similarity between $i$-th query and $j$-th target.

\begin{table*}[ht]
\centering
\begin{tabular}{c|l|cc|cc|cc|cc|ccc}
\hline
\multirow{2}{*}{Noise} & \multirow{2}{*}{Methods} 
& \multicolumn{2}{c|}{Dress} 
& \multicolumn{2}{c|}{Shirt} 
& \multicolumn{2}{c|}{Toptee} 
& \multicolumn{3}{c}{Average} \\
\cline{3-11}
& & R@10 & R@50 & R@10 & R@50 & R@10 & R@50 & R@10 & R@50 & AVG. \\
\hline
\hline
\multirow{8}{*}{0\%} 
& SSN~\cite{ssn}~(AAAI'24) & 34.36 & 60.78 & 38.13 & 61.83 & 44.26 & 69.05 & 38.92 & 63.89 & 51.40 \\
          & CALA~\cite{cala}~(SIGIR'24) & 42.38 & 66.08 & 46.76 & 68.16 & 50.93 & 73.42 & 46.69 & 69.22 & 57.96 \\
          & SPRC~\cite{sprc}~(ICLR'24) & 49.18 & 72.43 & 55.64 & 73.89 & \textbf{59.35} & \underline{78.58} & 54.72 & 74.97 & 64.85 \\
 & RCL~\cite{RCL}~(TPAMI'23) & 48.79 & 
          \underline{72.68} & 55.89 & 73.90 & 56.91 & 77.41 & 53.86 & 74.66 & 64.26 \\
          & RDE~\cite{rde}~(CVPR'24) & 47.84 & 71.89 & 54.37 & 73.55 & 56.91 & 77.21 & 53.04 & 74.22 & 63.63 \\
          & TME~\cite{TME}~(CVPR'25) & \underline{49.73} & 71.69 & \textbf{56.43} & \underline{74.44} & \underline{59.31} & \textbf{78.94} & \underline{55.16} & \underline{75.02} & \underline{65.09} \\
          
          & \textbf{INTENT (Ours)} & \textbf{50.32} & \textbf{72.10} & \underline{56.32} & \textbf{74.93} & {59.28} & {78.45} & \textbf{55.31} & \textbf{75.16} & \textbf{65.24} \\
          \cline{1-11}
\multirow{8}{*}{20\%}
& SSN~\cite{ssn}~(AAAI'24) & 22.61 & 45.56 & 27.87 & 48.58 & 31.82 & 55.28 & 27.43 & 49.81 & 38.62 \\
& CALA~\cite{cala}~(SIGIR'24) & 29.05 & 51.36 & 35.28 & 56.23 & 36.05 & 58.24 & 33.46 & 55.28 & 44.37 \\
& SPRC~\cite{sprc}~(ICLR'24) & 39.81 & 62.22 & 48.58 & 66.29 & 50.48 & 70.58 & 46.29 & 66.36 & 56.33 \\
& RCL~\cite{RCL}~(TPAMI'23) & 47.05 & \underline{70.65} & 53.14 & 71.74 & 55.28 & 75.62 & 51.82 & 72.67 & 62.25 \\
& RDE~\cite{rde}~(CVPR'24) & 44.62 & 68.91 & 50.74 & 69.09 & 52.12 & 73.38 & 49.16 & 70.64 & 59.81 \\
& TME~\cite{TME}~(CVPR'25)& \underline{49.03} & 70.35 & \textbf{55.84} & \underline{73.16} & \underline{57.22} & \underline{78.23} & \underline{54.03} & \underline{73.91} & \underline{63.97} \\
& \textbf{INTENT (Ours)} & \textbf{49.32}& \textbf{71.43} & \underline{55.32}& \textbf{73.57}& \textbf{58.01} & \textbf{78.46} & \textbf{54.22} & \textbf{74.49} & \textbf{64.36} \\
\hline
\multirow{8}{*}{50\%}
& SSN~\cite{ssn}~(AAAI'24) & 15.27 & 33.71 & 23.36 & 41.61 & 22.79 & 42.94 & 20.47 & 39.42 & 29.95 \\
& CALA~\cite{cala}~(SIGIR'24) & 20.77 & 40.95 & 29.69 & 46.57 & 27.03 & 46.81 & 24.83 & 44.78 & 34.80 \\
& SPRC~\cite{sprc}~(ICLR'24) & 35.94 & 57.16 & 42.25 & 61.63 & 44.98 & 64.76 & 41.06 & 61.19 & 51.12 \\
& RCL~\cite{RCL}~(TPAMI'23) & 43.68 & 66.44 & 50.74 & 69.19 & 52.63 & 73.84 & 49.01 & 69.82 & 59.42 \\
& RDE~\cite{rde}~(CVPR'24) & 41.30 & 64.75 & 47.06 & 66.34 & 50.13 & 70.63 & 46.16 & 67.24 & 56.70 \\
& TME~\cite{TME}~(CVPR'25)& \underline{46.26} & \underline{68.27} & \textbf{53.09} & \underline{71.88} & \underline{55.07} & \textbf{76.59} & \underline{51.47} & \underline{72.25} & \underline{61.86} \\
& \textbf{INTENT (Ours)} & \textbf{47.99} & \textbf{71.24} & \underline{52.78} & \textbf{72.48} & \textbf{56.79} & \underline{76.23} & \textbf{52.52} & \textbf{73.32} & \textbf{62.92} \\
\hline
\multirow{8}{*}{80\%}
& SSN~\cite{ssn}~(AAAI'24) & 11.16 & 25.24 & 16.98 & 30.72 & 17.03 & 32.64 & 15.05 & 29.53 & 22.29 \\
& CALA~\cite{cala}~(SIGIR'24) & 14.28 & 30.59 & 19.73 & 35.82 & 19.48 & 36.10 & 17.83 & 34.41 & 26.00 \\
& SPRC~\cite{sprc}~(ICLR'24) & 28.41 & 50.77 & 36.21 & 54.37 & 35.90 & 59.06 & 33.51 & 54.03 & 43.77 \\
& RCL~\cite{RCL}~(TPAMI'23) & 38.82 & 60.54 & 45.44 & 64.38 & 47.42 & 68.38 & 43.89 & 64.43 & 54.16 \\
& RDE~\cite{rde}~(CVPR'24) & 37.63 & 59.64 & 43.62 & 62.12 & 46.10 & 66.50 & 42.45 & 62.75 & 52.60 \\
& TME~\cite{TME}~(CVPR'25)& \underline{41.45} & \underline{64.35} & \underline{47.30} & \underline{68.20} & \underline{51.25} & \underline{73.23} & \underline{46.67} & \underline{68.60} & \underline{57.63} \\
& \textbf{INTENT (Ours)} & \textbf{42.07} & \textbf{65.58} & \textbf{50.38} & \textbf{69.41} & \textbf{53.09} & \textbf{73.91} & \textbf{48.51} & \textbf{69.63} & \textbf{59.07} \\
\hline
\end{tabular}
\caption{Performance comparison on the FashionIQ validation set in terms of R@K(\%). The best and second-best results are highlighted in \textbf{bold} and \underline{underline}, respectively.}
\label{tab:fiq_noise}
\end{table*}

We then perform dynamic decisions based on $p^+$ and $p^-$.
    \noindent\textbf{\textit{Negative Weight Reward}. }
    When all queries show low similarity to their positives within a batch, correct correspondences are more likely to exist among negatives. We thus apply Negative Weight Reward to negatives, as follows: 
    \begin{equation}
    \mathbf{N} =
    \{\textbf{n}_{ij}| \textbf{n}_{ij}=(1-p_i^+)\cdot(1 - y_{ij})\},
     \end{equation}
    where $\mathbf{n}_{ij}$ is the reward from $i$-th positive to $j$-th negative.
    
    \noindent\textbf{\textit{Positive Weight Reward}. }
    Complementarily, when all queries show low similarity to other negatives, positives are more likely to be cleanly labeled. We apply Positive Weight Reward to the positive group as follows:
    \begin{equation}
    \mathbf{R} =
    \{\textbf{r}_{ij}| \textbf{r}_{ij}=(1-p_i^-)\cdot y_{ij}\},
    \end{equation}
    where $\mathbf{r}_{ij}$ is the reward from $i$-th negative to $j$-th positive.
    
Finally, we construct a scalable decision boundary via both rewards, obtaining loyalty degree estimation matrix $\mathbf{L}$: 
    \begin{equation}
    \mathbf{L} = (\mathbf{S}+\mathbf{N}+\mathbf{R})/2 = 
    \{\textbf{l}_{ij}| \textbf{l}_{ij}= (\textbf{s}_{ij} + \textbf{n}_{ij} + \textbf{r}_{ij})/2\}.
    \label{eq:eq9}
    \end{equation}

Based on $\mathbf{L}$, we further propose the Soft Discriminative Loss, formulated as follows:
\begin{equation}
\mathcal{L}_{\text{sod}} = -\frac{1}{B}\sum_{i=1}^{B}\sum_{j=1}^{B} y_{ij}\log(\textbf{l}_{ij}),
\end{equation}
where $B$ is the batch size, $\textbf{l}_{ij}$ is the loyalty degree of the $i$-th query regarding the $j$-th candidate.

Finally, we obtain the final loss function of INTENT as,
    \begin{equation}
    \mathbf{\Theta^{*}}=
    \underset{\mathbf{\Theta}}{\arg \min } \left( {\mathcal{L}}_{robust}+\mu {\mathcal{L}}_{sod} +\alpha {\mathcal{L}}_{caco}
    \right),
    \label{eq:eq11}
    \end{equation}
where $\mathbf{\Theta^{*}}$ is the to-be-optimized parameter for INTENT and $\mu, \alpha$ are trade-off hyper-parameters.

\section{Experiments}
This section delves into our comprehensive experiments of INTENT and the corresponding analyses. Following the SOTA method TME~\cite{TME}, the noise ratio is set to 20\% for all ablation and parameter sensitivity experiments. 

\subsection{Experimental Settings}

\textbf{Datasets.}
Following previous works, we apply two standard datasets widely used in CIR for evaluation, including a fashion-domain dataset FashionIQ~\cite{FashionIQ}, and an open-domain dataset CIRR~\cite{cirr}.

\noindent
\textbf{Implementation Details.}
We adopt the BLIP-2~\cite{blip-2} backbone for INTENT. We set the number $Q$ of learned queries for the Q-former to $32$. And the embedding dimension $D$ is set to $256$.
Through a comprehensive grid search, we set $\mu=0.2, \alpha=0.6$ for both datasets.
We also adopt the temperature factor $\tau$ to $0.07$ for Eqn~$($\ref{robust}$)$.
We trained INTENT for 10 epochs using the AdamW optimizer with the initial learning rate of 4e-5, while the batch size is set to $128$ and the learning rate for CLIP is 1e-6. 
All experiments were conducted on a single NVIDIA A40 GPU.

\noindent
\textbf{Evaluation.}
We adopt the widely accepted metric Recall@K (short for R@K) to measure whether the target image is retrieved within top-K candidates. Following existing settings, for CIRR, we report overall recall at K=$1, 5, 10, 50$, and subset performance at K=$1, 2, 3$; for FashionIQ, we report for each of three categories at K=$10, 50$.

\subsection{Performance Comparison}
To assess the robustness of INTENT under NTC, we conduct experiments on CIRR and FashionIQ, comparing INTENT to both standard CIR models, and robust baselines. As shown in Table~\ref{tab:cirr-noise} and~\ref{tab:fiq_noise}, we further observe that:
\textbf{1) Robust methods outperform ordinary approaches.} According to Table~\ref{tab:cirr-noise} and~\ref{tab:fiq_noise}, across all noise ratios, robust methods such as RCL and TME consistently outperform conventional models like CALA and SPRC with this gap growing as noise increases. For example, on CIRR with a noise ratio of $20\%$, SPRC is only $2.76\%$ behind TME in Avg metric, and even outperforms some robust models on certain metrics. However, at $80\%$ noise, SPRC lags behind TME by $13.84\%$ in Avg. Similar results are observed on FashionIQ.
These results highlight the high sensitivity of conventional methods, while robust approaches maintain strong performance under severe noise.
\textbf{2) INTENT demonstrates superior robustness compared to other robust methods.} 
As Table~\ref{tab:fiq_noise} shows, on FashionIQ dataset, when noise ratio $\sigma=20\%$, INTENT surpasses TME by $0.39\%$ on Avg; as $\sigma$ increases to $50\%$ and $80\%$, this margin further widens to $1.06\%$ and $1.44\%$, respectively. 
Similar trends appear on CIRR. These results highlight INTENT's robustness in NTC scenarios.

\subsection{Ablation Study}
To assess the contribution of each component in INTENT, we conduct comprehensive ablation studies on both datasets.

\noindent\textbf{Ablation on modules and their components.}
As shown in Table~\ref{tab:module_ablation}, \textbf{A\#1} and \textbf{A\#2} denote variants where the VIC module is removed and where counterfactual images are replaced by grayscale images, respectively. \textbf{A\#3} - \textbf{\#5} correspond to removing $\mathbf{R}$, $\mathbf{N}$, or both from the loyalty degree matrix $\mathbf{L}$ in Eqn~(\ref{eq:eq9}).
Several key observations emerge from the results. 1) Both w/o VIC and w/o Intervention exhibit noticeable performance drops, demonstrating VIC's effectiveness in promoting visual invariance learning and reducing modality-inherent noise impact. The w/o Intervention still retains high performance, indicating the intervention process itself is robust and helps stabilize model training.
2) \textbf{A\#3}-\textbf{\#5}, all related to scalable decision boundaries, result in varying performance degradation. Notably, removing both rewards (w/o Both\_Reward) leads to the largest drop, highlighting scalable decision boundary's advantage over relying solely on raw similarity scores. 
Moreover, w/o NWR slightly outperforms w/o PWR, which is expected since with 20\% noise, most positives are real matches and thus benefit more from enhanced confidence in positive samples.
These results validate the contribution of BiODL in improving discrimination and robustness to correspondence noise.

\begin{table}[t!]
  \centering
  \tabcolsep=4pt
    \begin{tabular}{c|c|c|c|c|c}
    \hline
    \multicolumn{1}{c|}{\multirow{2}{*}{A\#}} & 
    \multicolumn{1}{c|}{\multirow{2}{*}{Derivative}} & 
    \multicolumn{2}{c|}{FashionIQ-Avg} &
    \multicolumn{2}{c}{CIRR-Avg} \\
    \cline{3-6}  & & \multicolumn{1}{c}{R@10}  & R@50 & \multicolumn{1}{c}{R@K}   & R$_{sub}$@K \\
    \hline
    \hline
    \multicolumn{6}{c}{\textit{\textbf{Visual Invariant Composition (VIC)}}} \\
    \hline
    \multicolumn{1}{c|}{ 1} & \multicolumn{1}{l|}{w/o VIC} & \multicolumn{1}{c}{53.05} & \multicolumn{1}{c|}{72.98} & \multicolumn{1}{c}{79.36} & 86.12 \\
    \multicolumn{1}{c|}{2} & \multicolumn{1}{l|}{w/o  Intervention} & \multicolumn{1}{c}{54.01} & \multicolumn{1}{c|}{73.67} & \multicolumn{1}{c}{80.07} & 87.52 \\
    \hline
    \multicolumn{6}{c}{\textit{\textbf{Bi-Objective Discriminative Learning (BiODL)}}} \\
    \hline
    \multicolumn{1}{c|}{3} & \multicolumn{1}{l|}{w/o PWR} & \multicolumn{1}{c}{53.37} & \multicolumn{1}{c|}{73.97} & \multicolumn{1}{c}{79.80} & 87.28 \\
    \multicolumn{1}{c|}{4} & \multicolumn{1}{l|}{w/o NWR} & \multicolumn{1}{c}{53.89} & \multicolumn{1}{c|}{74.07} & \multicolumn{1}{c}{80.03} & 87.33 \\
    \multicolumn{1}{c|}{5} & \multicolumn{1}{l|}{w/o Both\_Raward} & \multicolumn{1}{c}{52.48} & \multicolumn{1}{c|}{73.20} & \multicolumn{1}{c}{77.93} & 85.86 \\
    \hline
         \multicolumn{2}{c|}{\textbf{INTENT(Ours)}}  & \multicolumn{1}{c}{\textbf{54.22}} & \textbf{74.49} & \multicolumn{1}{c}{\textbf{80.17}} & \textbf{88.60} \\
    \hline
    \end{tabular}%
  \caption{The ablation study for \textbf{modules} of INTENT.}
  \label{tab:module_ablation}
  \vspace{-12pt}
\end{table}

\noindent \textbf{Ablation on loss functions.}
As shown in Table~\ref{tab:loss_ablation}, \textbf{B\#1} and \textbf{B\#3} ablate the Robust Contrastive Loss and Soft Discriminative Loss in BiODL, while \textbf{B\#2} applies $\mathcal{L}_{\text{robust}}$ without masking positives.
\textbf{B\#4} – \textbf{B\#6} replace the CKA metric in VIC’s Causal Consistency Loss with MSE, L1, or L2, respectively.
We have the following observations. 1) Both w/o robust and w/o sod cause notable performance drops, confirming their complementary roles in discriminating noisy and real correspondence.
2) The largest decline occurs with w/o mask. This is expected since removing the mask causes the model to push positive samples apart, severely degrading performance.
3) $\mathcal{L}_{\text{caco}}$ w/ MSE, L1, L2 all lead to a performance decline compared to CKA. This may be due to CKA specifically measures similarities between the relational structures of two features (i.e., centered Gram matrices), rather than merely minimizing point-wise differences.

\begin{table}[t!]
\tabcolsep=4pt
  \centering
    \begin{tabular}{c|c|c|c|c|c}
    \hline
    \multicolumn{1}{c|}{\multirow{2}{*}{B\#}} & 
    \multicolumn{1}{c|}{\multirow{2}{*}{Derivative}} & 
    \multicolumn{2}{c|}{FashionIQ-Avg} & 
    \multicolumn{2}{c}{CIRR-Avg} \\
    \cline{3-6}  \multicolumn{1}{c|}{} & \multicolumn{1}{c|}{} & \multicolumn{1}{c}{R@10}  & \multicolumn{1}{c|}{R@50} & \multicolumn{1}{c}{R@K}   & R$_{sub}$@K \\
    \hline
    \hline
    \multicolumn{6}{c}{\textit{\textbf{Loss Functions}}} \\
    \hline
     1    & \multicolumn{1}{l|}{w/o robust} & \multicolumn{1}{c}{51.14} & \multicolumn{1}{c|}{72.25} & \multicolumn{1}{c}{78.33} & 86.06 \\
     2    &  \multicolumn{1}{l|}{w/o mask} & \multicolumn{1}{c}{50.46} & \multicolumn{1}{c|}{70.87} & \multicolumn{1}{c}{74.26} & 82.88 \\
     3     & \multicolumn{1}{l|}{w/o sod} & \multicolumn{1}{c}{51.06} & \multicolumn{1}{c|}{72.07} & \multicolumn{1}{c}{77.67} & 84.65 \\
     \hline
     4     & \multicolumn{1}{l|}{$\mathcal{L}_{\text{caco}}$ w/ MSE} & \multicolumn{1}{c}{53.88} & \multicolumn{1}{c|}{73.77} & \multicolumn{1}{c}{79.69} & 87.71 \\
     5     & \multicolumn{1}{l|}{$\mathcal{L}_{\text{caco}}$ w/ L1} & \multicolumn{1}{c}{53.99} & \multicolumn{1}{c|}{74.15} & \multicolumn{1}{c}{79.92} & 87.88 \\
     6     & \multicolumn{1}{l|}{$\mathcal{L}_{\text{caco}}$ w/ L2} & \multicolumn{1}{c}{53.79} & \multicolumn{1}{c|}{73.90} & \multicolumn{1}{c}{79.79} & 88.01 \\
    \hline
          \multicolumn{2}{c|}{\textbf{INTENT(Ours)}} & \multicolumn{1}{c}{\textbf{54.22}} & \textbf{74.49} & \multicolumn{1}{c}{\textbf{80.17}} & \textbf{88.60} \\
    \hline
    \end{tabular}%
  \caption{The ablation study for \textbf{loss functions} of INTENT.}
    \label{tab:loss_ablation}%
\end{table}%

\begin{figure}[t]
\begin{center}
\includegraphics[width=0.95\linewidth]{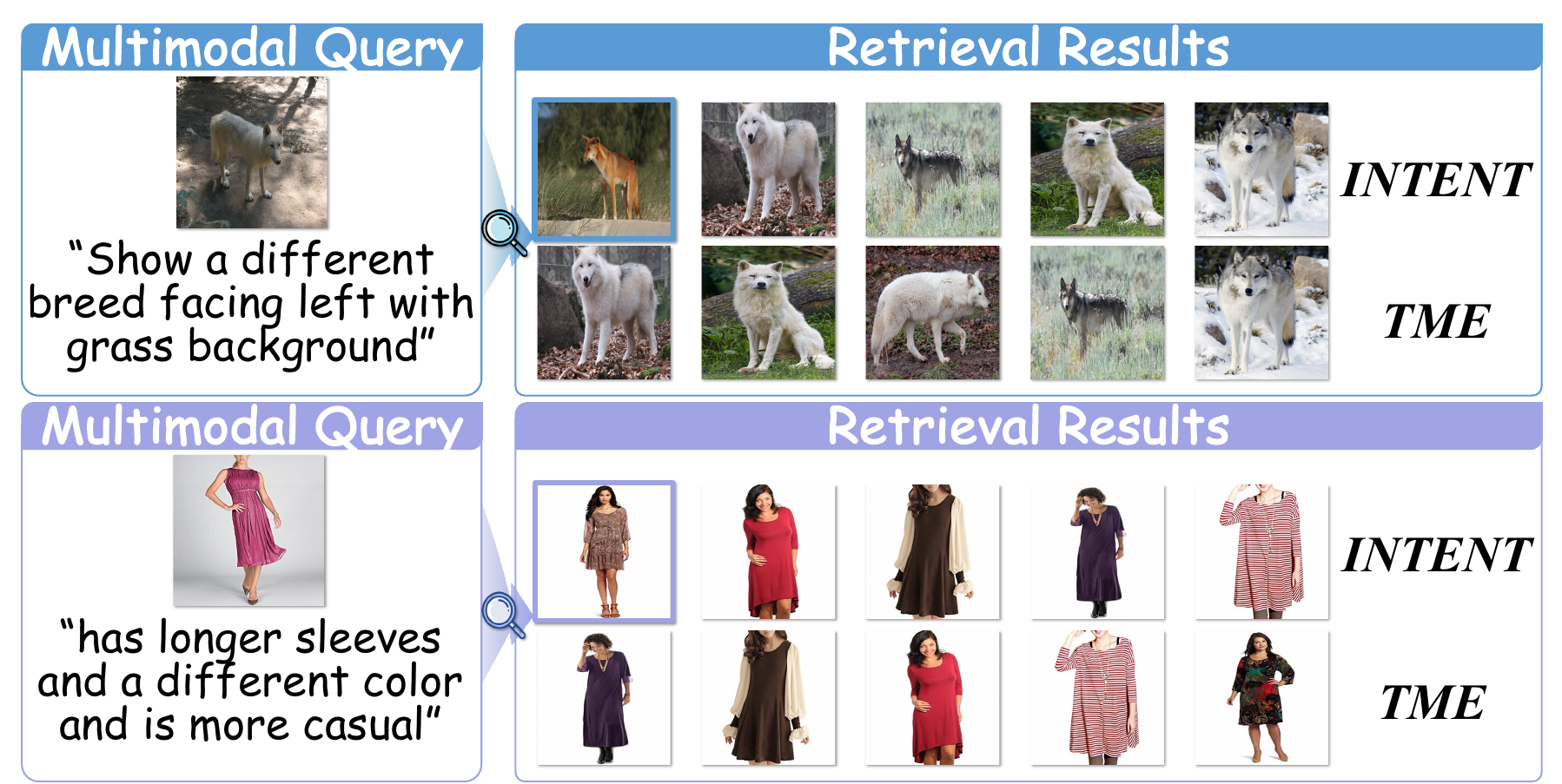}
\end{center}
   \caption{Case Study on (a) CIRR and (b) FashionIQ.}
\label{fig:case}
\vspace{-12pt}
\end{figure}

\subsection{Case Study}
To intuitively demonstrate the accuracy and robustness of INTENT, we compare its top-5 retrieval results with those of the sub-optimal robust method TME on FashionIQ and CIRR, as shown in Figure~\ref{fig:case}, where images in colored boxes are target images.
INTENT successfully retrieves the target images at top-1 by capturing causal relationships during multimodal query composition and effectively ignoring potential modality-inherent noise.
However, TME fails in these cases and does not retrieve the target even within the top-5 results. 
We attribute this to TME's lack of addressing noise prior to composition, and utlizing hard decision boundary to discriminate correspondences, without addressing modality-inherent noise or employing decision optimization.

\section{Conclusion}
In this work, we investigate neglected inherent noise and hard decision boundaries in NTC. To address these issues, we propose INTENT, comprising two components: Visual Invariant Composition and Bi-Objective Discriminative Learning. 
The former uses causal intervention via FFT to generate intervened composed features, promoting visual invariance and allowing the model to ignore modality-inherent noise during composition.
The latter employs collaborative optimization with positive/ negative samples, establishing a scalable decision boundary that dynamically adjusts decisions according to the loyalty degree, enabling robust correspondence discrimination. Extensive experiments on two benchmarks reveal superiority and robustness of INTENT.

\section*{Acknowledgments}
This work was supported in part by the National Natural Science Foundation of China, No.:62276155, No.:62576195, and No.:62572282
\clearpage
\appendix
\appendix
\noindent
This is the supplementary material of the submitted paper \textit{\textbf{``INTENT: Invariance and Discrimination-aware Noise Mitigation for Robust Composed Image Retrieval''}}. The content catalog is as follows:
\begin{itemize}
    \item \textbf{Appendix~\ref{appendix:validity}}: Module Validity Analysis
    \begin{itemize}
        \item \textbf{Appendix~\ref{appendix:validity_vic}}: Visual Analyses and Ablation Studies on Frequency-Domain Intervention in the VIC Module

        \item \textbf{Appendix~\ref{appendix:validity_biodl}}: Visual Analysis of the Scalable Decision Boundary in the BiODL Module
    \end{itemize}
    \item \textbf{Appendix~\ref{appendix:datasets}}: Datasets
        \item \textbf{Appendix~\ref{appendix:additional_performance_efficiency}}: Efficiency Evaluation
       \item    
\textbf{Appendix~\ref{appendix:algorithm}}: Algorithm of INTENT's Training Procedure

    \item \textbf{Appendix~\ref{appendix:qualitative}}: More Qualitative Results
    \begin{itemize}
        \item \textbf{Appendix~\ref{appendix:qualitative_matrix}}: Similarity Matrix
        \item \textbf{Appendix~\ref{appendix:qualitative_more_case}}: More Case Study
    \end{itemize}
\end{itemize} 

\section{Module Validity Analysis}
\label{appendix:validity}
\subsection{Visual Analyses and Ablation Studies on Frequency-Domain Intervention in the VIC Module}
\label{appendix:validity_vic}
\begin{figure}[t]
    \centering
    \includegraphics[width=1.00\linewidth]{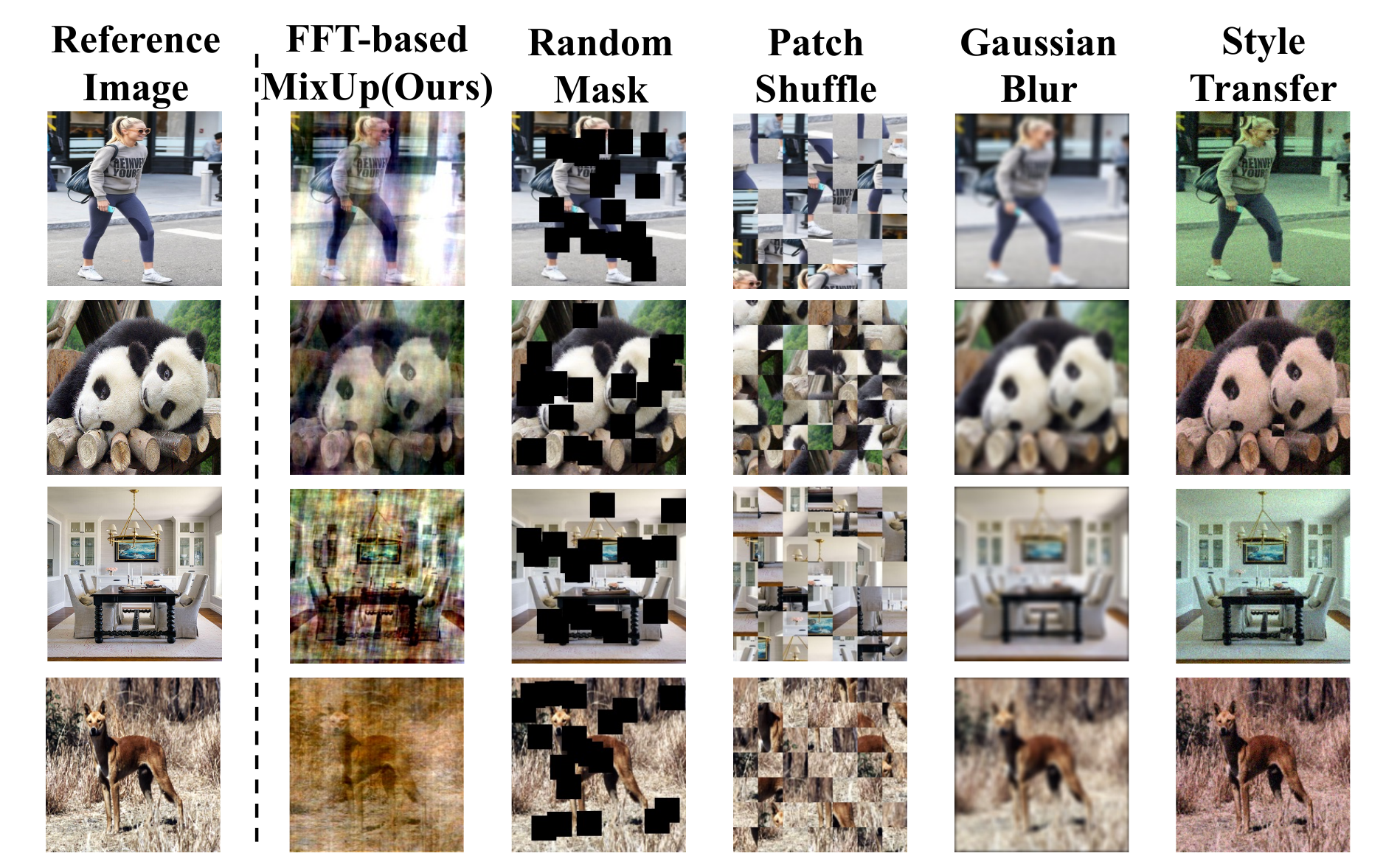} 
    \caption{Visualization of the effects of different intervention operations in the VIC module, including our FFT-based intervention, Random Mask, Patch Shuffle, Gaussian Blur, and Style Transfer.}
    \label{fig:Visual in the VIC Module}
\end{figure}

Figure~\ref{fig:Visual in the VIC Module} provides visual examples of five intervention operations which can be taken in VIC module, including FFT-based MixUp (\textbf{ours}), Random Mask, Patch Shuffle, Gaussian Blur, and Style Transfer. We have reported FFT-based MixUp in our full paper. As the figure shows, \textbf{Random Mask} randomly occludes patches of the image, producing blocky artifacts and breaking spatial continuity. \textbf{Patch Shuffle} disrupts local coherence by shuffling image patches, causing significant fragmentation. \textbf{Gaussian Blur} uniformly smooths the image, removing both high-frequency noise and some semantic details. \textbf{Style Transfer} applies the style of a randomly selected image onto the reference image, resulting in subtle but consistent changes in texture and color while largely preserving object structure.
The corresponding quantitative results on the CIRR dataset (Table~\ref{tab:intervention}) further reveal the impact of each operation on retrieval performance under a $20\%$ noise ratio.

From the visualizations, FFT-based MixUp and Style Transfer both yield counterfactual images that maintain global semantic structures and style, while introducing subtle but effective changes to the appearance. In contrast, Random Mask and Patch Shuffle severely disrupt the spatial integrity of the image, leading to visually fragmented samples. Gaussian Blur, meanwhile, indiscriminately removes high-frequency details, often at the cost of critical semantic cues.

These observations are echoed in the performance comparison. The bottom row of Table~\ref{tab:intervention} shows that FFT-based MixUp achieves the best results across nearly all evaluation metrics, with an average (R@$5$, Rsubset@$1$) of $\textbf{79.66}$, outperforming all other intervention methods. Style Transfer also delivers relatively strong performance ($\textbf{78.24}$), confirming its effectiveness in generating semantically meaningful counterfactuals. However, Random Mask, Patch Shuffle, and Gaussian Blur not only fail to improve the model, but also lead to substantial performance drops (with averages as low as $\textbf{71.75}$), indicating that excessive or unsupervised corruption can be detrimental to model training.

We attribute the advantage of FFT-based MixUp over other methods, including Style Transfer, to its ability to perturb high-frequency noise in a globally consistent yet semantically aligned manner, effectively challenging the model's invariance to modality-inherent noise without destroying essential visual structures. In contrast, the first three operations (Random Mask, Patch Shuffle, Gaussian Blur) either destroy local continuity or blur away key semantics, thus harming model generalization and robustness.

\begin{table*}[t]
  \tabcolsep=2pt
  \centering
    \begin{tabular}{rc|cccc|ccc|c}
    \hline
    \multicolumn{1}{c}{\multirow{2}[2]{*}{C\#}} 
    & \multirow{2}[2]{*}{Operation} 
    & \multicolumn{4}{c|}{R@K}  & \multicolumn{3}{c|}{Rsubset@K} & \multirow{2}[2]{*}{Avg(R@5,Rsubset@1)} \\
    \multicolumn{1}{c}{} &       & K=1   & K=5   & K=10  & K=50  & K=1   & K=2   & K=3   &  \\
    \hline
    \hline
    \multicolumn{1}{l}{1} & \multicolumn{1}{l|}{Random Mask} & 46.62 & 77.60 & 85.90 & 93.54 & 71.79 & 86.21 & 92.10 & 74.70 \\
    \multicolumn{1}{l}{2} & \multicolumn{1}{l|}{Patch Shuffle} & 43.73 & 74.33 & 82.58 & 90.97 & 69.17 & 85.57 & 91.08 & 71.75 \\
    \multicolumn{1}{l}{3} & \multicolumn{1}{l|}{Gaussian Blur} & 48.60 & 79.79 & 87.85 & 95.76 & 73.76 & 89.14 & 95.13 & 76.78\\
    \multicolumn{1}{l}{4} & \multicolumn{1}{l|}{Style Transfer} & 50.02 & 80.90 & 89.47 & 97.66 & 75.57 & 90.12 & 96.05 & 78.24 \\
    \hline
          \multicolumn{2}{l|}{\textbf{FFT-based MixUp}} & \textbf{51.25} & \textbf{81.36} & \textbf{90.02} & \textbf{98.05} & \textbf{77.95} & \textbf{91.40} & \textbf{96.46} & \textbf{79.66} \\
    \hline
    \end{tabular}%
\caption{Performance comparison of interventions using different operations in VIC module on CIRR. Noise ratio is set to $20\%$.}
  \label{tab:intervention}%
\end{table*}%

\subsection{Visual Analysis of the Scalable Decision Boundary in the BiODL Module}
\label{appendix:validity_biodl}
\begin{figure}[t]
\begin{center}
\includegraphics[width=1.00\linewidth]{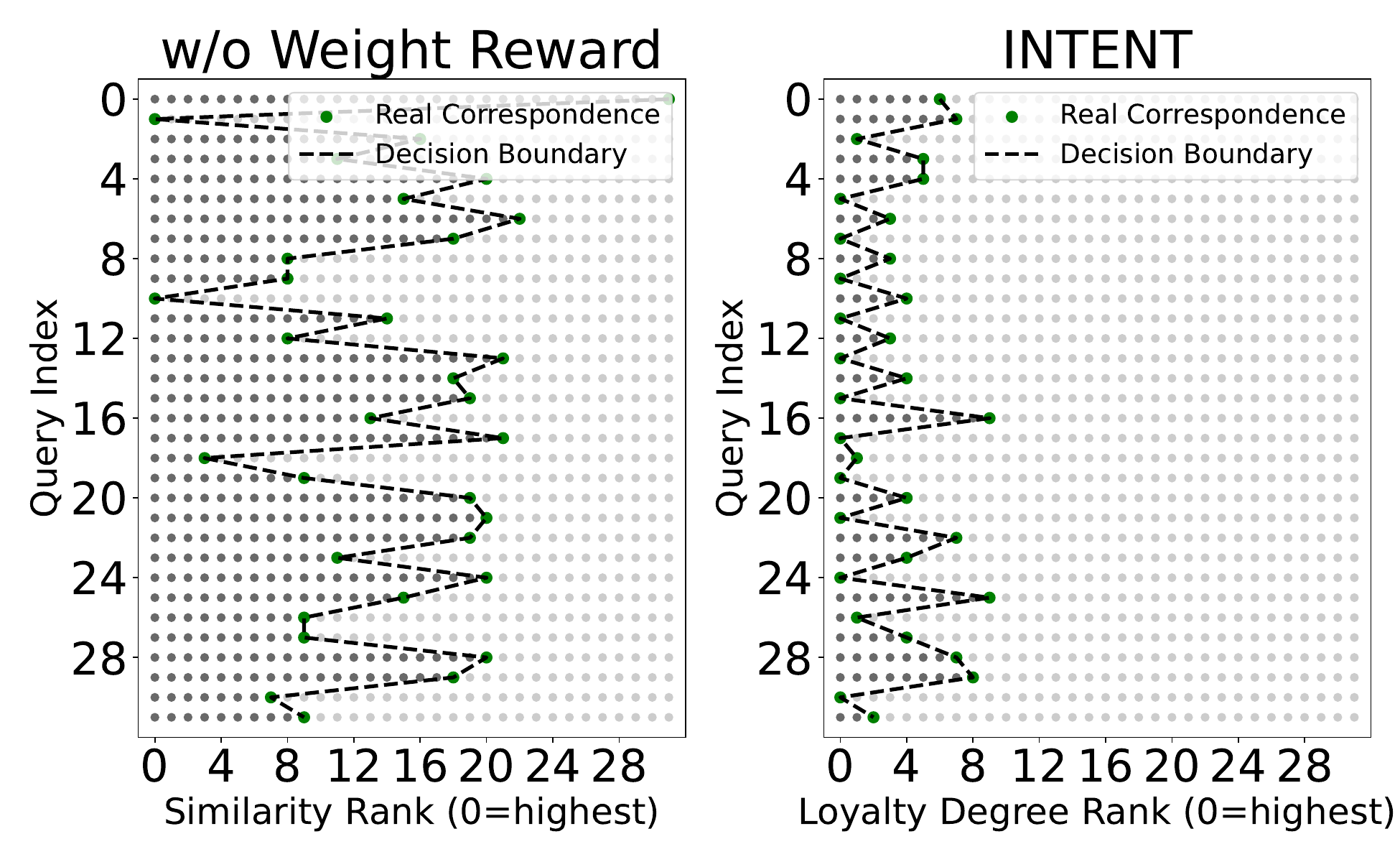}
\end{center}
   \caption{Similarity (left) and loyalty degree (right) distribution of each query with respect to candidate targets within a batch.
The vertical axis represents the indices of individual queries, while the horizontal axis denotes the indices of candidate targets for each query, ordered by their similarity /  (left) or loyalty degree (right), from highest to lowest. A smaller index along the horizontal axis thus indicates a candidate with a higher similarity / loyalty degree for the corresponding query, with index 0 representing the candidate exhibiting the highest similarity / loyalty degree.
The green dots indicate the real correspondence (i.e., the true matching target) for each query, while the gray dots correspond to all other candidate samples. By connecting the green dots across queries, one obtains an approximate visualization of our decision boundary.}
\label{fig:decision_boundary}
\end{figure}

Figure~\ref{fig:decision_boundary} illustrates the distribution of similarity (left) and loyalty degree (right) ranks for each query with respect to candidate targets within a batch, comparing the INTENT variant w/o Weight Reward (left) and the full INTENT model (right). 
The vertical axis indicates query indices, while the horizontal axis represents the candidate targets ranked by loyalty degree, with smaller indices denoting higher loyalty degrees. 
Green dots correspond to the real correspondence (ground-truth positive), and the dashed line connects these, approximating the decision boundary learned by the model.
In the w/o Weight Reward setting, the distribution of green dots is dispersed, with many real correspondences failing to rank among the top similarity positions. 
Instead, they are often located toward the middle or even right side of the rank axis, indicating that the model struggles to consistently assign high confidence to the true matching pairs. 
This results in a non-smooth, unstable decision boundary, making the model more susceptible to confusion between real correspondences and noisy samples.

By contrast, the full INTENT model (right) demonstrates a significant improvement: the green dots cluster much closer to the left side, with most real correspondences being ranked highest in loyalty degree for their respective queries. This shows that the model, empowered by the scalable decision boundary strategy, is able to assign high confidence to genuine matches and systematically push noisy or irrelevant candidates to the lower confidence region (right side, gray dots). The learned boundary is therefore more robust and better aligned with the true correspondence distribution.

These results strongly validate the effectiveness of our scalable decision boundary in BiODL module. The Weight Reward strategy allows INTENT to dynamically adjust the decision boundary based on the loyalty degree distribution, ensuring more stable and reliable identification of real correspondence, even under varying semantic complexity or noise ratios. This dynamic adaptation not only boosts the utilization of real correspondences for optimization, but also effectively suppresses the impact of hard negatives and noisy samples.
The visualization provides clear evidence that the scalable decision boundary realized via the Weight Reward mechanism plays a crucial role in enhancing both the discrimination of real correspondences and the robustness against cross-modal correspondence noise. 
This underpins the superior performance and noise resistance demonstrated by INTENT in challenging NTC scenarios.

\section{Datasets}
\label{appendix:datasets}
To evaluate the effectiveness of the INTENT model, we selected two widely used benchmark datasets for the composed image retrieval task. 
\begin{itemize}
    \item \textbf{FashionIQ Dataset.} FashionIQ~\cite{FashionIQ} is a dataset specifically designed for fashion-oriented image retrieval. It consists of $77,684$ online images and contains $30,134$ annotated triplets, covering three representative categories of fashion items: dresses, shirts, and T-shirts.
    \item \textbf{CIRR Dataset.} CIRR~\cite{cirr} is constructed based on real-world scene images originating from the NLVR2 natural language visual reasoning dataset. CIRR comprises $36,554$ annotated triplets and $21,552$ images. 
\end{itemize}
Compared to FashionIQ, CIRR emphasizes complex interactions among multiple objects in natural scenes, thereby alleviating the limitations of overfitting to a specific domain. Furthermore, CIRR addresses the issue of frequent hard negatives in FashionIQ by including fine-grained contrastive subsets and an incomplete label design, enabling more robust evaluation under challenging real-world conditions.


\begin{table*}[h]
  \tabcolsep=1.0pt
  \centering
    \begin{tabular}{c|c|c|c|c|c|c|c|c}
    \hline
    Type  & Method & FLOPs & Parameters & GPU Memory & Test time & Train Time & FashionIQ-Avg & CIRR-Avg \\
    \hline
    \hline
    Ordinary & SPRC  & 413.38G & 915.69 M & 24478MiB(bs=128) & 0.011s/sample & 2.624s/iteration & 56.33 & 76.98 \\
    \hline
    \multirow{3}{*}{Robust Methods} & TME   & 405.2G & 915.68M & 12405MB(bs=128) & \multicolumn{1}{p{6em}|}{0.124s/sample} & 7.858s/iteration & 63.97 & \textbf{79.74} \\
    \cline{2-9}          & w/o VIC & 605.13G & 915.89M & 16873MB(bs=128) & \multicolumn{1}{p{6em}|}{0.0097s/sample} & 2.564s/iteration & 63.02 & 77.93 \\
    \cline{2-9}          & Ours  & 605.13G & 915.89M & 23405MB(bs=128) & 0.010s/sample & 3.195s/iteration & \textbf{64.36} & 79.66 \\
    \hline
    \end{tabular}%
\caption{Comparison of computational complexity and efficiency among SPRC, TME, and INTENT with its ablation variants.}
  \label{tab:efficiency}%
\end{table*}%

\section{Efficiency Evaluation}
\label{appendix:additional_performance_efficiency}
As shown in Table~\ref{tab:efficiency}, FLOPs indicates the computational complexity per single inference process; 
Test Time refers to the average time required to process a single query during inference; 
Train Time measures the time cost and corresponding GPU resource consumption per training epoch;
GPU Memory and Parameters represent the memory usage during training and the model's parameter count, respectively;
and Avg (Recall) quantifies the average retrieval performance on the CIRR and FashionIQ datasets.

In terms of efficiency, our proposed INTENT model demonstrates outstanding operational efficiency and resource utilization while maintaining a similar parameter size.
Although the FLOPs of INTENT are slightly higher than those of TME, the inference time per sample is significantly reduced to just $0.010$ seconds, roughly $1/12$ that of TME, indicating substantially improved single-sample inference efficiency.
During training, INTENT requires only $3.195$ seconds per iteration, compared to $7.858$ seconds for TME, resulting in a $59\%$ increase in training efficiency and demonstrating a notable acceleration in computational speed.

Additionally, we analyzed the efficiency of the ablation variant (w/o VIC). 
This variant maintains nearly the same inference speed as the full model ($0.0097$ s/sample) and achieves even faster training ($2.564$ s/iteration), but this comes at the cost of lower retrieval performance. Specifically, the average recall on both FashionIQ and CIRR decreases compared to full INTENT, and even falls below TME, highlighting the crucial role of the VIC module in enhancing causal representation and semantic discrimination.

INTENT exhibits remarkable advantages in both computational efficiency and retrieval performance, greatly enhancing robustness against noisy correspondence in CIR task. 
The results demonstrate that the VIC module is pivotal for the model's causality and overall effectiveness, ensuring both superior accuracy and operational efficiency.

\section{Algorithm of INTENT's Training Procedure}
\label{appendix:algorithm}
As shown in Algorithm~\ref{alg:algorithm}, the training algorithm of INTENT integrates frequency-domain intervention, multimodal composition, and robust optimization in a unified pipeline. 
The process begins by generating a counterfactual view of the reference image through amplitude perturbation in the frequency domain using FFT, ensuring that only the modality-inherent noise is altered while preserving semantic structure. 
Both the original and counterfactual images are then composed with the modification text using the Q-Former, yielding paired composed features. 
By comparing the feature representations of these pairs with a CKA-based causal consistency loss, INTENT explicitly enforces the invariance of semantic content under visual perturbation, allowing the model to ignore irrelevant noise during feature composition.
Subsequently, the model learns to distinguish true correspondences from noisy triplets via two complementary objectives. 
A robust contrastive loss focuses on negative sample discrimination, while a soft discriminative loss, based on an adaptively constructed loyalty degree matrix, provides a scalable decision boundary for sample matching. 
The overall training objective combines these losses with the causal consistency constraint, enabling INTENT to jointly suppress modality-inherent noise and cross-modal correspondence noise. 
This pipeline ensures that INTENT achieves robust and accurate retrieval, even in the presence of complex noise and annotation ambiguity.


\begin{algorithm}[t!]
\caption{Algorithm of INTENT's Training Procedure}
\label{alg:algorithm}
\textbf{Input}: Reference image \( x_r \), target image \( t \), modification text \( y_m .\) \\
\textbf{Parameter}: Soft Discriminative Loss weight $\mu$, Causal Consistency Loss weight $\alpha$. \\
\textbf{Output}: Fine-tuned model. \\
\begin{algorithmic}[1]
\REQUIRE Reference image $x_r$, modification text $y_m$, target image $t$, irrelevant image $x_d$ (sampled), batch size $B$, temperature $\tau$, trade-off parameters $\mu, \alpha$

\STATE {// \textbf{1. Counterfactual Image Generation via FFT Intervention}}
\STATE Apply FFT to $x_r$ and $x_d$ to obtain frequency spectra: $F_r = FFT(x_r)$, $F_d = FFT(x_d)$
\STATE Mix amplitude spectra in central frequency region with random ratio $\lambda$ to obtain intervened amplitude: $\hat{A}_r = \lambda A_{\text{crop}}^d + (1 - \lambda) A_{\text{crop}}^r$
\STATE Reconstruct counterfactual image $\hat{x}_r$ with inverse FFT using original phase: $\hat{x}_r = FFT^{-1}(\hat{A}_r, \theta_r)$

\STATE {// \textbf{2. Multimodal Fusion with Q-Former}}
\STATE Obtain composed features from original and counterfactual images:
\\ $F_c = \mathrm{Q\text{-}Former}(\Phi_X(x_r), \Phi_Y(y_m))$
\\ $\hat{F}_c = \mathrm{Q\text{-}Former}(\Phi_X(\hat{x}_r), \Phi_Y(y_m))$

\STATE {// \textbf{3. Causal Consistency Loss (CKA)}}
\STATE Compute Gram matrices $K_c$, $L_c$ for $F_c$, $\hat{F}_c$ and perform centering to obtain $\bar{K}_c$, $\bar{L}_c$
\STATE Calculate Causal Consistency Loss:
\\ $\mathcal{L}_{\text{caco}} = \frac{1}{B} \sum_{i=1}^B \left( 1 - \frac{\langle \bar{K}_c^i, \bar{L}_c^i \rangle_F}{\| \bar{K}_c^i \|_F \| \bar{L}_c^i \|_F} \right)$

\STATE {// \textbf{4. Robust Contrastive Loss for Negative Pairs}}
\STATE For all $i, j$ in batch, calculate robust InfoNCE loss focusing on negatives:
\\ $\mathcal{L}_{\text{robust}} = - \frac{1}{B} \sum_{i \neq j} \log \left( 1 - \frac{\exp (F_c^i F_t^j / \tau)}{\sum_{j=1}^B \exp (F_c^i F_t^j / \tau)} \right)$

\STATE {// \textbf{5. Loyalty Degree Matrix Construction for Adaptive Decision Boundary}}
\STATE Compute similarity matrix $S$ between all composed queries $F_c$ and targets $F_t$
\STATE For each query $i$, compute $p_i^+$ (max positive matching) and $p_i^-$ (max negative matching)
\STATE Construct Negative Weight Reward $N$ and Positive Weight Reward $R$:
\\ $N = \{\ n_{ij} = (1 - p_{i}^{+})\cdot(1 - y_{ij}) \}$
\\ $R = \{\ r_{ij} = (1-p_i^-)\cdot y_{ij}\}$
\STATE Compute loyalty degree matrix: $L = (S+N+R)/2$
\STATE Calculate soft discriminative loss:
\\ $\mathcal{L}_{\text{sod}} = - \frac{1}{B} \sum_{i=1}^B \sum_{j=1}^B y_{ij} \log (l_{ij})$

\STATE {// \textbf{6. Joint Optimization}}
\STATE Update model parameters by minimizing total loss:
\\ $\Theta^* = \arg \min_{\Theta} \left( \mathcal{L}_{\text{robust}} + \mu \mathcal{L}_{\text{sod}} + \alpha \mathcal{L}_{\text{caco}} \right)$
\end{algorithmic}
\end{algorithm}

\section{More Qualitative Results}
\label{appendix:qualitative}

\subsection{Similarity Matrix}
\label{appendix:qualitative_matrix}
\begin{figure}[h]
    \centering
    \includegraphics[width=1.00\linewidth]{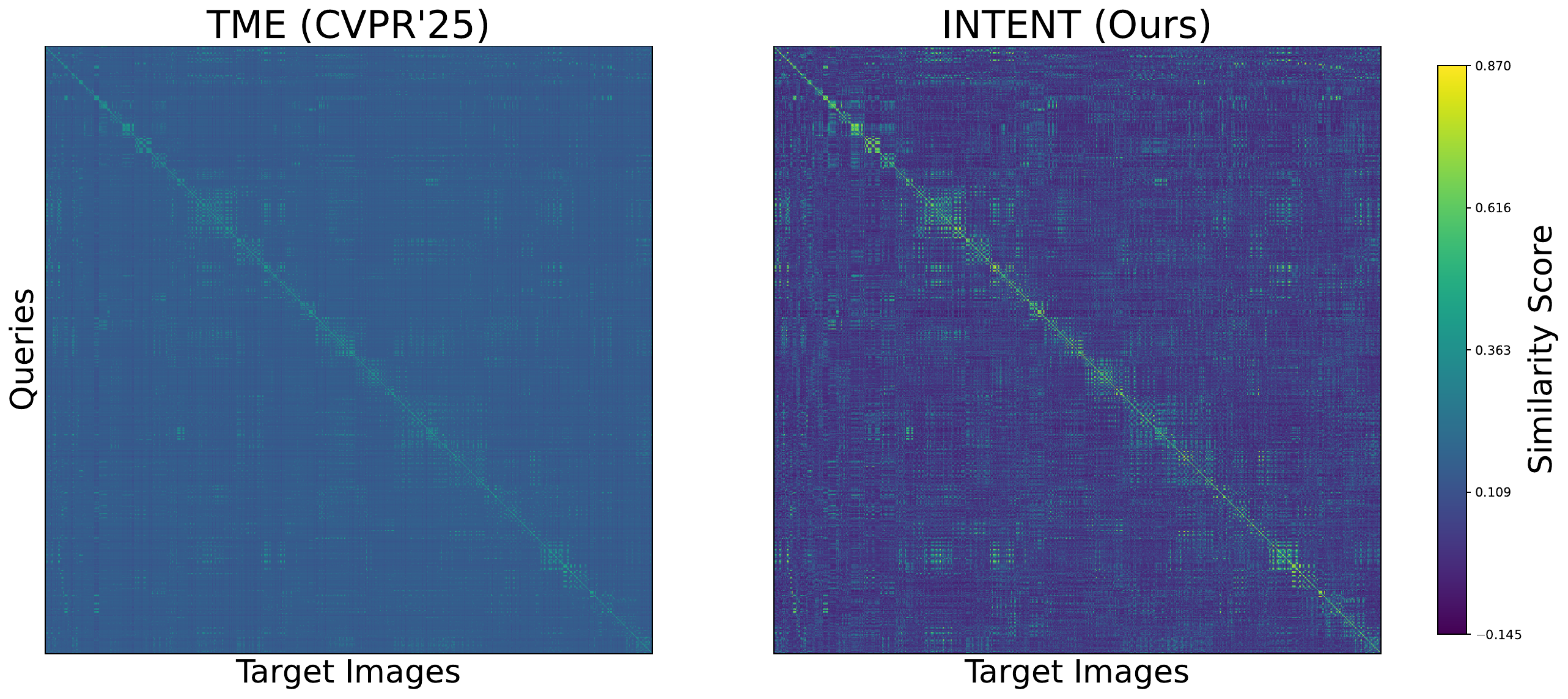} 
    \caption{Comparison of similarity matrices between TME (CVPR'25) and INTENT (Ours) on the CIRR validation set. The matrices visualize the pairwise similarity scores between all queries and target images. Brighter diagonal elements indicate stronger correspondence between matched pairs.}
    \label{fig:sim}
\end{figure}

As shown in Figure~\ref{fig:sim}, the similarity matrix produced by INTENT exhibits a more pronounced and continuous diagonal, with clearer contrast between positive matches (diagonal) and non-matching pairs (off-diagonal), compared to TME. 
This indicates that INTENT more effectively aligns true correspondences while suppressing spurious similarities. The result highlights the contribution of the Bi-Objective Discriminative Learning (BiODL) module, which dynamically constructs scalable decision boundaries and adapts to noisy or correct correspondence scenarios. Consequently, INTENT demonstrates superior discrimination and robustness, ensuring that positives are accurately identified and negatives are effectively suppressed.

\subsection{More Case Study}
\label{appendix:qualitative_more_case}
\begin{figure*}[ht]
    \centering
\includegraphics[width=0.8\linewidth]{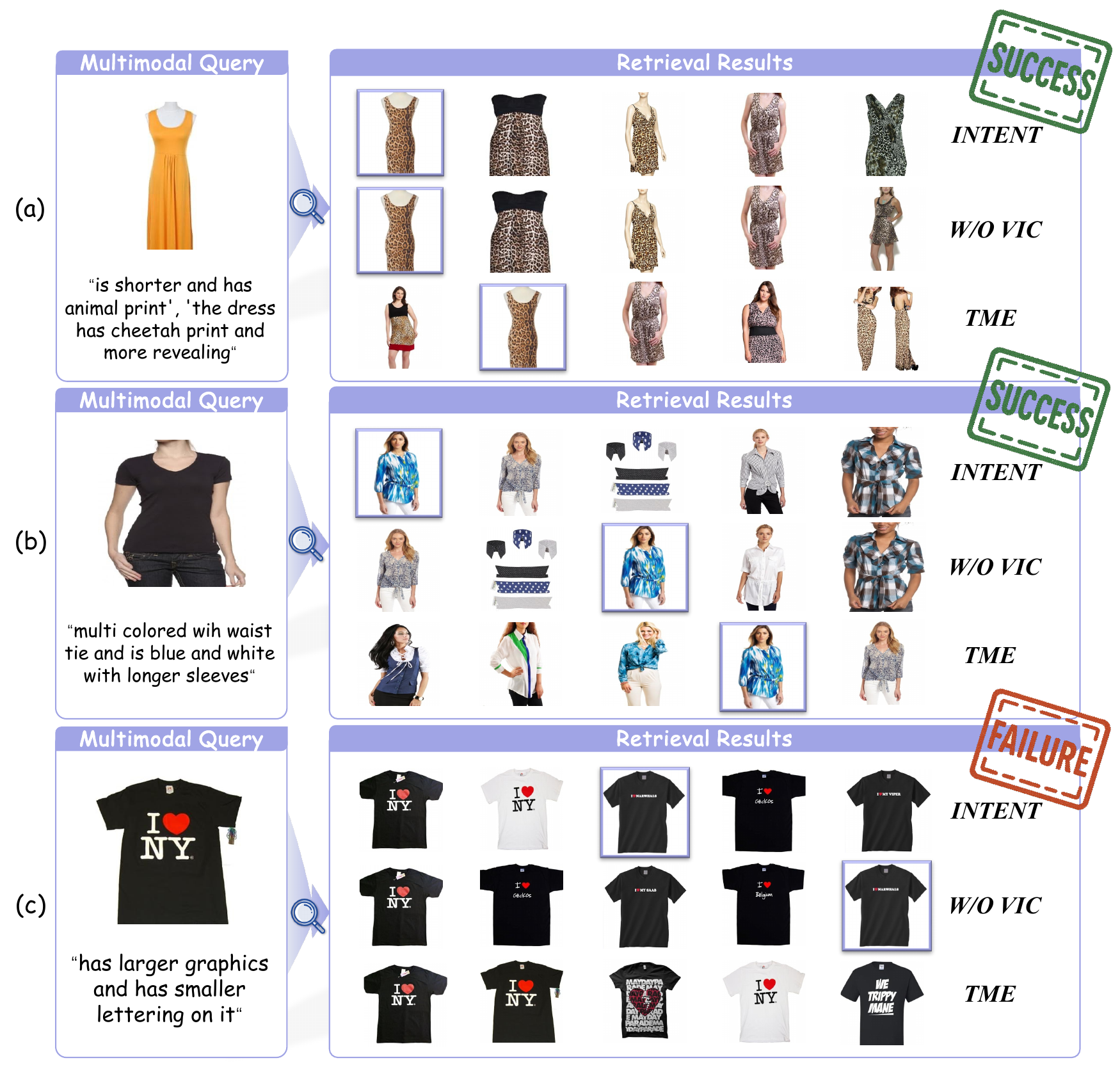} 
    \caption{Case study on FashionIQ dataset.}
    \label{fig:case_fashioniq_faliure}
\end{figure*}

\begin{figure*}[ht]
    \centering
    \includegraphics[width=0.8\linewidth]{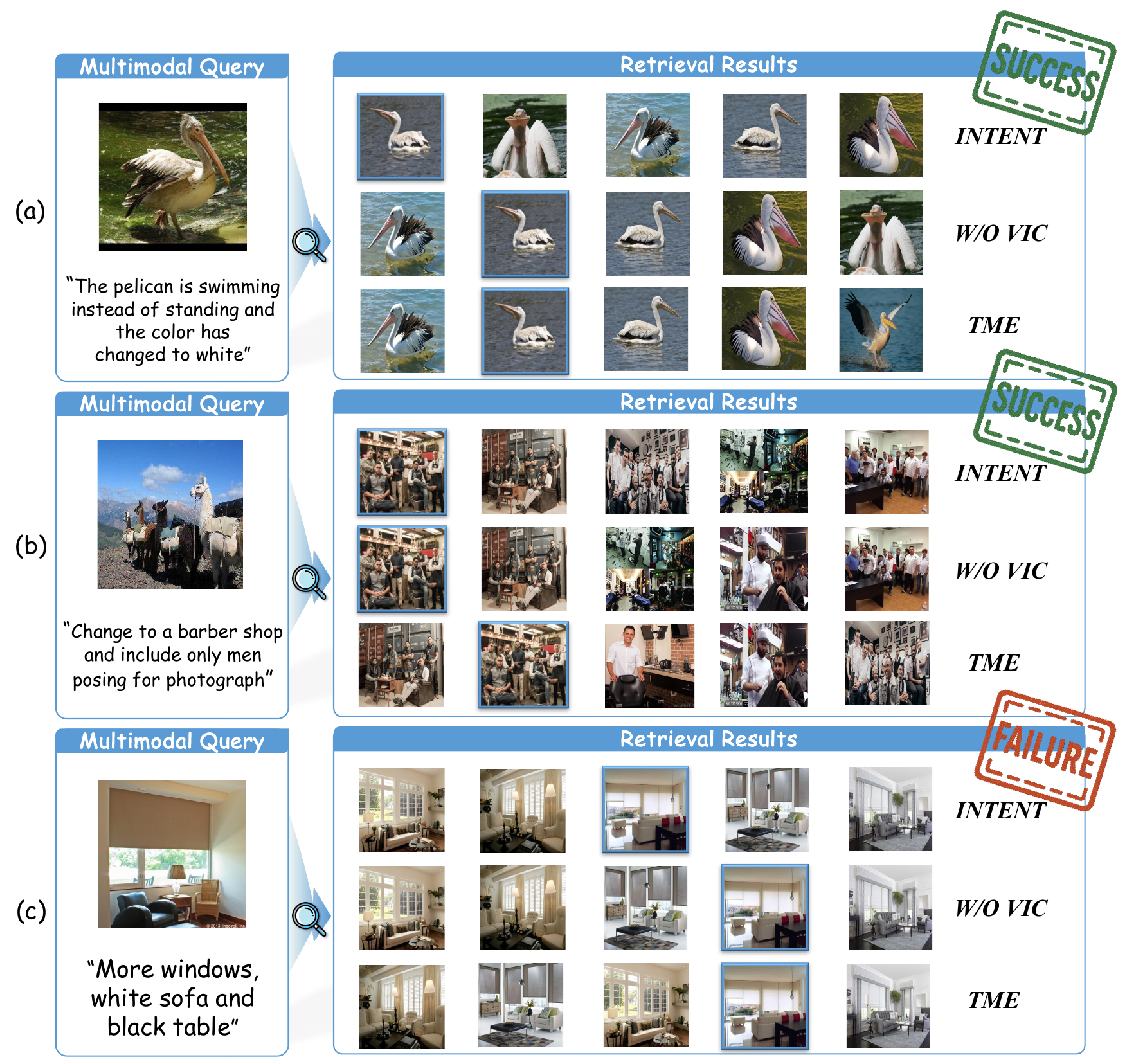} 
    \caption{Case study on CIRR dataset.}
    \label{fig:case_cirr_faliure}
\end{figure*}

To further validate the effectiveness of the INTENT in the CIR task, we conducted a visual case study on two standard datasets, FashionIQ and CIRR.
Figure~\ref{fig:case_fashioniq_faliure} and \ref{fig:case_cirr_faliure} present qualitative retrieval results for INTENT, its ablation variant (w/o VIC), and the representative robust method TME, showing the top-K (K=5) retrieved target images for each multimodal query. Each case consists of a reference image and a modification text. The image with a colored border is the ground-truth target, and its ranking directly reflects the accuracy of the retrieval method.

In the three FashionIQ retrieval examples in Figure~\ref{fig:case_fashioniq_faliure}, INTENT consistently ranks the correct target image in the top-5 in all scenarios, demonstrating precise semantic alignment with the multimodal query. 
For instance, in Figure~\ref{fig:case_fashioniq_faliure}(b), where the modification text is ``multi colored with waist tie and is blue and white with longer sleeves", INTENT successfully retrieves the target image at the top-1 position. In contrast, TME fails to identify the key characteristics required for accurate retrieval, resulting in evident semantic mismatches in its results.
The ablation variant w/o VIC, while able to retrieve some semantically relevant images and occasionally approaches the full INTENT in certain cases, generally produces inferior ranking.
This highlights the critical role of the VIC module: by introducing causal intervention, it enables the model to addressing modality-inherent noise, thus enhancing semantic alignment.

Similarly, the CIRR cases in Figure~\ref{fig:case_cirr_faliure} demonstrate INTENT's clear advantage in noise mitigation under complex modification scenarios involving animals, indoor scenes, and crowds.
INTENT not only consistently retrieves the correct target image at top-1, but also ensures that its top-ranked results conform closely to the modification requirements, further validating its dynamic adjustment of the scalable decision boundary between positive and negative samples in challenging contexts.

These qualitative cases comprehensively showcase INTENT's superiority in CIR, with significant improvements in accuracy over TME and its own ablation variant. 
They also illustrate the model's robustness to handle complex semantic modifications. This provides further evidence for the effectiveness of the proposed causal consistency learning and discriminative optimization strategies based on loyalty degree in addressing semantic uncertainty and NTC in CIR tasks.

\subsubsection{Failure Cases.} 
We also observed several failure cases for our model, such as those shown in Figure~\ref{fig:case_fashioniq_faliure}(c) and Figure~\ref{fig:case_cirr_faliure}(c). 
In these examples, neither INTENT, its ablation variant (w/o VIC), nor the baseline TME was able to accurately retrieve the target image at the top-1 position. 
However, it is worth highlighting that INTENT not only successfully includes the ground-truth target among the top-ranked candidates, but the other images it retrieves are also highly consistent with the query semantics in terms of visual attributes, even though these images are not labeled as "target images" in the dataset. 
We attribute this phenomenon to the potential presence of false negative samples within the datasets.

By comparison, the w/o VIC variant exhibits a decline in overall retrieval performance but still outperforms TME, whose top-5 results for this shirt query are less relevant to the modification. 
Although this scenario appears to constitute a “failure case,” it actually underscores, from another perspective, the critical value of the proposed causal consistency learning for complex semantic composition tasks.


\small
\bibliography{aaai2026}
\end{document}